\documentclass[11pt,authoryear,preprint]{elsarticle}
\usepackage[latin9]{inputenc}
\usepackage{geometry}
\usepackage{amsmath}
\usepackage{amssymb}
\usepackage{graphicx}

\makeatletter

\providecommand{\tabularnewline}{\\}

\usepackage{caption}
\usepackage{color}
\usepackage{lscape}
\usepackage{afterpage}
\usepackage{pstricks}
\usepackage{pst-plot}
\usepackage{longtable}
\usepackage{dcolumn}
\usepackage{pst-node}
\usepackage{amsthm}
\usepackage{multirow}
\usepackage{colortab}
\usepackage{color}
\usepackage{array}
\usepackage{slashbox}
\usepackage{colortbl}
\usepackage{subfigure}
\usepackage{shadbox}
\usepackage{tabularx}
\usepackage{textcomp}
\usepackage{pstricks}
\usepackage{pst-node}
\usepackage{pst-all}
\usepackage{algorithm}
\usepackage{algorithmic}\usepackage{url}
\psset{arrows=->, labelsep=3pt, mnode=circle}

\usepackage{eurosym}

\usepackage{hyperref}

\usepackage{amsthm,amsmath,eurosym}

\usepackage{etoolbox}
\makeatletter
\patchcmd{\ps@pprintTitle}{\footnotesize\itshape
Preprint submitted to \ifx\@journal\@empty Elsevier
\else\@journal\fi\hfill\today}{\relax}{}{}

\makeatother

\begin{document}

\begin{frontmatter}{}

\title{A robust hierarchical nominal classification method based on similarity
and dissimilarity using loss function and an improved version of the
deck of cards method}

\author[ceg-ist,inesc-id]{Ana Sara Costa}

\ead{anasaracosta@tecnico.ulisboa.pt}

\author[Eco]{Salvatore Corrente}

\ead{salvatore.corrente@unict.it}

\author[Eco,por]{Salvatore Greco}

\ead{salgreco@unict.it}

\author[ceg-ist]{Jos{'e} Rui Figueira}

\ead{figueira@tecnico.ulisboa.pt}

\author[inesc-id]{Jos{'e} Borbinha}

\ead{jlb@tecnico.ulisboa.pt}

\address[ceg-ist]{CEG-IST, Instituto Superior T{'e}cnico, Universidade de Lisboa, Portugal}

\address[inesc-id]{INESC-ID, Instituto Superior T{'e}cnico, Universidade de Lisboa, Portugal}

\address[Eco]{Department of Economics and Business, University of Catania, Italy}

\address[por]{CORL, Portsmouth Business School, University of Portsmouth, United
Kingdom}
\begin{abstract}
\noindent \textsc{Cat-SD} (\textsc{Cat}egorization by Similarity-Dissimilarity)
is a multiple criteria decision aiding method for dealing with nominal
classification problems (predefined and non-ordered categories). Actions
are assessed according to multiple criteria and assigned to one or
more categories. A set of reference actions is used for defining each
category. The assignment of an action to a given category depends
on the comparison of the action to each reference set according to
likeness thresholds. Distinct sets of criteria weights, interaction
coefficients, and likeness thresholds can be defined per category.
When applying \textsc{Cat-SD} to complex decision problems, may be
useful to consider a hierarchy of criteria helping to give a more
intelligible vision of the performances of the considered actions.
We propose to apply Multiple Criteria Hierarchy Process (MCHP) to
\textsc{Cat-SD}. An adapted MCHP is proposed to take into account
possible interaction effects between criteria structured in a hierarchical
way. On the basis of the known deck of cards method, we also consider
an imprecise elicitation of parameters permitting to take into account
interactions and antagonistic effects between criteria. The elicitation
procedure we are proposing can be applied to any \textsc{Electre}
method. With the purpose of exploring the assignments obtained by
\textsc{Cat-SD} considering possible sets of parameters, we propose
to apply the Stochastic Multicriteria Acceptability Analysis (SMAA).
The SMAA methodology allows to draw statistical conclusions on the
classification of the actions. The proposed method, SMAA-h\textsc{Cat-SD},
helps the decision maker to check the effects of the variation of
parameters on the classification at different levels of the hierarchy.
We propose also a procedure, based on the concept of loss function,
to obtain a final classification fulfilling some requirements given
by the decision maker and taking into account the hierarchy of criteria
and the probabilistic assignments obtained applying SMAA. Also this
procedure can be applied to any classification \textsc{Electre} method.
The application of the new proposal is shown through an example. 
\end{abstract}
\end{frontmatter}{}

\section{Introduction}

\noindent \label{Intro}

\noindent In several decision situations, we face a classification
problem involving the assessment of a set of actions (or alternatives),
according to multiple criteria (usually conflicting), and their assignment
to categories defined in a nominal way. In fact, the wide range of
potential real-world applications in various areas (e.g., human resources
management, finance, medicine, etc.) has motivated researchers to
develop Multiple Criteria Decision Aiding (MCDA) methods for dealing
with multiple criteria nominal classification problems. In this kind
of classification problems, categories are predefined and no order
exists among them (nominal categories). In opposition, in sorting
problems (or ordinal classification problems), there is a preference
order among the categories. In other fields, such as statistics and
machine learning (ML), both terms discrimination and classification
are used to refer to decision problems where the categories are defined
a priori and there is no preferential order among them. The term s\textit{upervised
classification problems} is usually used when the categories are previously
defined, whereas \textit{unsupervised classification problems} is
used when there is no information about the categories and they are
identified a posteriori (they are designed clusters) \citep{henriet2000thesis,Perny1998}.
In clustering, the objective is to find such clusters, representing
groups of actions with similar features. Recent proposals for handling
classification problems are mainly based on operations research and
artificial intelligence techniques \citep{Doumpos2002,sorting_review}.
In fact, nominal classification has been addressed in an MCDA setting,
but also in ML. The main difference of the MCDA setting from the standard
nominal classification problems in ML is the role of criteria. Standard
ML algorithms assume features (usually called attributes), whereas
MCDA assumes criteria. In particular, criteria in MCDA have, in general,
an increasing or a decreasing direction of preference that reveal
the preferences of the Decision Maker (DM) on such criteria. On the
contrary, features in ML have not any direction of preference and,
instead, the relation between the values of the attributes and the
preferences of the DM are discovered from data \citep{CGKSml}.

In the literature, we can find proposals for nominal classification
mainly using outranking-based procedures \citep{belacel2000multicriteria,henriet2000thesis,Leger2002,Perny1998,rigopoulos2010nexclass},
rough set theory \citep{slowinski2000generalized}, and verbal decision
analysis \citep{furems2013dominance}. The majority of existing MCDA
nominal classification methods are based on outranking relations (see,
for example, \citealt{belacel2000multicriteria,Perny1998}). While
for choice, ranking and sorting problems outranking binary relations
are acceptable, for nominal classification problems, they may be questionable.
One may argue that in nominal classification the aim of the pairwise
comparison should be to know whether the two actions are similar and
not if one action is preferred to the other one. None of the current
methods proposed a way to model preference information related to
similarity concepts when comparing actions, neither to deal with criteria
hierarchy and interactions between criteria. In addition, robustness
concerns have not been considered, and it has been pointed out as
an important issue in nominal classification \citep{sorting_review}.
The \textsc{Cat-SD} (\textsc{Cat}egorization by Similarity-Dissimilarity)
method has been recently proposed as a new MCDA method, covering some
of these issues \citep{Costaetal2018}. This method allows to assign
actions to nominal categories, based on similarity and dissimilarity
between actions, using reference actions to define the categories.
Multiple criteria and possible interactions in some pairs of criteria
are considered. In \textsc{Cat-SD}, for each category, a particular
set of preference parameters can be chosen (e.g., criteria weights
and interaction coefficients), which means that distinct parameter
sets can be defined among categories. Thus, \textsc{Cat-SD} has been
designed to model subjective judgments of the DM in pairwise comparison
of actions in terms of similarity and dissimilarity between the two
actions. Then, likeness binary relations are constructed taking into
account the preferences of the DM. Moreover, to the best of our knowledge,
\textsc{Cat-SD} is the first MCDA nominal classification method that
permits to model interactions between criteria. As stated in \citet{Costaetal2018},
there are still aspects that need further research related to \textsc{Cat-SD,
}namely considering a hierarchical structure of criteria and robustness
analysis, while different vectors of parameter sets are taken into
consideration.

In several decision aiding scenarios, complex multiple criteria decision
problems arise involving a great number of criteria for assessing
actions \citep{Belton,GreFigEhr,Ishi_nem}. The heterogeneity and
the high number of criteria are the main reasons for the complexity
of the decision problems. Structuring the criteria in a hierarchical
way can be a useful approach for dealing with such decision problems.
Multiple Criteria Hierarchy Process (MCHP) has been proposed to handle
the decision problems in which the considered criteria are hierarchically
structured \citep{CGShierarchy,CGSoutr,CGSElectreTri,Corrente2017}.
MCHP imposes a hierarchical structure of criteria, which means that
all criteria are not considered at the same level, and criteria are
grouped into subsets according to distinct points of view. In this
way, the elicitation of preferences of the DM related to the criteria
can be easier than considering a great number of heterogeneous criteria
at the same level. Indeed, MCHP has been applied, for example, to
a ranking method, \textsc{ Electre} III \citep{Electre_CFGS}, and
to sorting methods, such as\textsc{ Electre} \textsc{Tri}-B, \textsc{Tri}-C
and \textsc{Tri}-nC \citep{CGSElectreTri}, and \textsc{Promethee}
methods \citep{CGSoutr}. MCHP has also been applied to the aggregation
of interacting criteria by means of the Choquet integral in \citet{angilella2016MCHP}.
To the best of our knowledge, there is no research work adopting such
an approach to multiple criteria nominal classification methods. In
this paper, we propose to apply MCHP to the \textsc{Cat-SD} method.

We introduce an adapted MCHP to handle the three types of interaction
between criteria considered in \textsc{Cat-SD}: mutual-strengthening
effect, mutual-weakening effect, and antagonistic effect (for more
details on the meaning of these effects in case of outranking relations,
see \citealp{figueira2009electre}). Moreover, an imprecise elicitation
of criteria weights is considered. For that, we adopt an extension
of the Simos-Roy-Figueira (SRF) \citep{FigueiraRo02} by considering
imprecise preference information provided by the DM to assign values
to the criteria weights \citep{Electre_CFGS}. To take into account
interaction between criteria, we further extend this methodology obtaining
a new version that can be applied to any \textsc{Electre} method considering
such interaction.

The assignment results provided by the \textsc{Cat-SD} method can
include multiple assignments of an action, i.e., a given action can
be assigned to several categories. It is interesting to know the robustness
of the assignment of each action, considering then the robustness
of the recommendations with respect to the assignment results. In
this sense, to take into account all sets of weights and interaction
coefficients compatible with the information provided by the DM, we
propose to apply the Stochastic Multicriteria Acceptability Analysis
(SMAA) \citep{Lahdelma,Lahdelma_book_greco,pelissari2019smaa} to
draw conclusions with respect to the assignments of each action. Application
of SMAA allows to obtain, for each action, the probability of its
assignment to each category (or a set of categories), not only when
considering the whole set of criteria, but also when considering a
particular node in the hierarchy structure. Of course, this can be
a relevant information for the DM. Since, finally, one classification
has to be selected, we propose also a methodology permitting to define
a single classification taking into consideration the whole probabilistic
information related to the imprecise elicitation of preference parameters.
The procedure we propose is based on the concept of loss function
and it has an autonomous interest that permits to apply this approach
to any classification method, not only nominal but also ordinal.

Our aim is therefore to present a new method, in the sense of a more
comprehensive framework, for dealing with these interrelated issues
by adopting an integrated approach. Thus, we can take advantage of
the main characteristics of the methods that we propose to integrate:
\begin{itemize}
\item[$-$] \textit{Hierarchy of criteria:} The use of the MCHP is beneficial
for the user from two perspectives. On one hand, the DM can provide
information not only at comprehensive level but considering a particular
aspect of the problem. Indeed, it can be a bit upset in comparing
two actions considering all criteria simultaneously but the DM can
feel more confident in expressing the preferences taking into account
one or some aspects he knows more. On the other hand, the DM can get
information not only at global level but also at partial one, and
this is an added value since the DM can discover the weak and strong
points of the actions at hand;
\item[$-$] \textit{The imprecise SRF method:} Asking the DM to provide exact
values for all the parameters involved in the model is meaningless
even for one expert in MCDA. In general, the DM is more confident
in exercising the preferences than in justifying them. For this reason,
we use the imprecise SRF method to obtain the criteria weights by
asking the DM to provide preference information in an imprecise way;
\item[$-$] \textit{SMAA:} The motivations on the basis of the use of SMAA are
strictly connected to the previous point. Indeed, in general, more
than one set of values of parameters can be compatible with the information
provided by the DM and choosing only one or some of them to get the
final recommendations on the problem at hand can be considered arbitrary
to some extent. A recommendation built taking into account the plurality
of preference parameters compatible with the information provided
by the DM is more robust and, consequently, more trustworthy;
\item[$-$] \textit{Robustness concerns:} The DM is interested in a final recommendation
that takes into account the robustness concerns represented by the
results of SMAA. Therefore, as already mentioned, we propose a procedure
that, starting from the probability to be classified to different
categories supplied by SMAA, provides a comprehensive classification
fulfilling some possible requirements given by the DM. 
\end{itemize}
The main objectives of this paper are (leading to a more general framework): 
\begin{enumerate}
\item To apply MCHP to the nominal classification method \textsc{Cat-SD}; 
\item To use the imprecise SRF method for each category taking into account
the hierarchy of criteria and the possible interactions between criteria;
the method we propose has a general interest and can be applied to
any outranking method considering interactions between criteria; 
\item To apply SMAA to the hierarchical \textsc{Cat-SD} method by sampling
several sets of parameters compatible with some preferences provided
by the DM; 
\item To propose a procedure that starting from the probabilistic assignments
obtained by SMAA provides a final classification that fulfills some
requirements given by the DM; the method we propose has a general
interest and can be applied to any classification method, both nominal
and ordinal. 
\end{enumerate}

It is worth to remark that the parameters elicitation is a fundamental
step not only for our method but for all methods using an indirect
preference information provided by the DM. The weights elicitation
as well as the interaction coefficients elicitation can involve a
certain difficulty and different methods have been proposed in literature
to this aim. For instance, \citet{FigueiraRo02} provides a method
to elicit the weights and \citet{figueira2009electre} presents an
elicitation procedure for getting the values of the interaction effects
(see also \citealt{bottero2015dealing} and \citealt{CostaChapterART}
applying such a procedure to real-world cases).

This paper is organized as follows. Section \ref{CAT-SD} introduces
the \textsc{Cat-SD} method. Section \ref{MCHP_CAT_SD} is related
to our proposal of applying MCHP to the \textsc{Cat-SD} method, in
order to construct the hierarchical \textsc{Cat-SD} method, h\textsc{Cat-SD}.
Section \ref{ImpreciseSRF} presents a way for dealing with imprecise
information to determine the criteria weights when considering the
h\textsc{Cat-SD} method. Section \ref{SMAASec} is devoted to the
application of SMAA to the h\textsc{Cat-SD} method, building the comprehensive
method SMAA\nobreakdash-h\textsc{Cat-SD}. Section \ref{sec:RobustAssignment}
proposes a procedure to obtain the final nominal classification results
according to some requirements indicated by the DM. Section \ref{Example}
provides a numerical example of application of the SMAA\nobreakdash-\hspace{0pt}h\textsc{Cat-SD}
method. Section \ref{Conclusions} presents some concluding remarks
and future lines of research.


\section{The CAT-SD method}

\noindent \label{CAT-SD}
In this section, we briefly introduce the \textsc{Cat-SD} method (for
more details, see \citealt{Costaetal2018}). This method deals with
decision problems where categories are defined in a nominal way (they
are not ordered). Each category is defined a priori and characterized
by a set of reference actions. Each action is assessed on several
criteria, and assigned to a category or a set of categories. The assignment
of actions is based on the concepts of similarity and dissimilarity
between two actions. The main notation, concepts and definitions are
presented.

\subsection{Main notation }

\noindent In the \textsc{Cat-SD} method, the following notation is
used:
\begin{itemize}
\item[$-$] $A=\{a,...,a_{i}...\}$ is the set of actions (or alternatives) not
necessarily known a priori;
\item[$-$] $G=\{g_{1},...,g_{j},...,g_{n}\}$ is the set of all criteria\footnote{In the following, for the sake of simplicity and without loss of generality
we shall write $g_{j}\in G$ or $j\in G$ interchangeably.};
\item[$-$] $C=\{C_{1},...,C_{h},...,C_{q},C_{q+1}\}$ is the set of nominal
categories, where $C_{q+1}$ is a dummy one considered to receive
actions not assigned to the other categories;
\item[$-$] $B=\{B_{1},...,B_{h},...,B_{q+1}\}$ is the set of all reference
actions, where \textbf{$B_{q+1}=\ensuremath{\emptyset}$};
\item[$-$] $\{b_{h1},\ldots,b_{h\ell},...,b_{h\vert B_{h}\vert}\}$ is the set
of (representative) reference actions chosen to define category $C_{h}$,
for $h=1,...,q$;
\item[$-$] $k_{j}^{h}$ is the weight of criterion $g_{j}$ for category $C_{h}$,
for $j=1,...,n$ and $h=1,...,q$;
\item[$-$] $k_{j\ell}^{h}$ is a mutual-strengthening (or mutual-weakening)
coefficient of the pair of criteria $\{g_{j},g_{\ell}\}$, with $k_{j\ell}^{h}>0$
(or $k_{j\ell}^{h}<0$), for $h=1,...,q$;
\item[$-$] $k_{j|p}^{h}$ is an antagonistic coefficient for the ordered pair
of criteria $(g_{j},g_{p})$, with $k_{j|p}^{h}<0$, for $h=1,...,q$;
\item[$-$] $k(C_{h})$ is the set of all criteria weights and interaction coefficients
of category $C_{h}$, for $h=1,...,q$;
\item[$-$] $\lambda^{h}$ is a likeness threshold of category $C_{h}$, for
$h=1,...,q$.
\end{itemize}

\subsection{Modeling similarity-dissimilarity }

\noindent \textsc{Cat-SD} is more focused on similarity between actions
than on their dissimilarity, since likeness between actions is usually
what counts most when categorizing actions. According to a given criterion,
when an action $a$ (the subject) is compared to an action $b$ (the
referent or the reference action), similarity-dissimilarity between
them can be assessed. Indeed, the preferences of the DM with respect
to the similarity\nobreakdash-dissimilarity between the two actions
on a criterion can be modeled through a function.

In what follows, let $E_{j}$ denote the scale of criterion $g_{j}$,
$j=1,...,n$ (generally bounded from below by $g_{j}^{\min}$ and
from above by $g_{j}^{\max}$). Consider the difference of performances
of actions $a$ and $b$, $\Delta_{j}(a,b)=diff\left\{ g_{j}(a),g_{j}(b)\right\} $.
Let $E_{\Delta_{j}}$ denote the scale of such a difference. For ratio
and interval scales, $diff\left\{ g_{j}(a),g_{j}(b)\right\} =g_{j}(a)-g_{j}(b),$
and for ordinal scales, $diff\left\{ g_{j}(a),g_{j}(b)\right\} $
corresponds to the number of performance levels between $g_{j}(a)$
and $g_{j}(b)$. Without loss of generality, we assume that criteria
are to be maximized. A \textit{per}\emph{\nobreakdash-\hspace{0pt}criterion
similarity\nobreakdash-\hspace{0pt}dissimilarity function} is a
real-valued function $f_{j}\;:\;E_{\Delta_{j}}\rightarrow[-1,1]$
such that: 
\begin{enumerate}
\item $f_{j}$ is a non-decreasing function of $\Delta_{j}(a,b)$, if $\Delta_{j}(a,b)\in[-diff\{g_{j}^{\max},g_{j}^{\min}\},0]$; 
\item $f_{j}$ is a non-increasing function of $\Delta_{j}(a,b)$, if $\Delta_{j}(a,b)\in[0,diff\{g_{j}^{\max},g_{j}^{\min}\}]$; 
\item $f_{j}>0$ iff criterion $g_{j}$ contributes to similarity; 
\item $f_{j}<0$ iff criterion $g_{j}$ contributes to dissimilarity. 
\end{enumerate}
\noindent This function defines:
\begin{itemize}
\item[$-$] A \emph{per\nobreakdash-\hspace{0pt}criterion similarity function}
$s_{j}(a,b)=f_{j}\big(\Delta_{j}(a,b)\big)$, if $f_{j}\big(\Delta_{j}(a,b)\big)>0$,
and $s_{j}(a,b)=0$, otherwise;
\item[$-$] A\emph{ per\nobreakdash-\hspace{0pt}criterion dissimilarity function}
$d_{j}(a,b)=f_{j}\big(\Delta_{j}(a,b)\big)$, if $f_{j}\big(\Delta_{j}(a,b)\big)<0$,
and $d_{j}(a,b)=0$, otherwise.
\end{itemize}
\noindent The parameters of a function $f_{j}$ can be induced with
the following set of questions for the DM (possibly supported by the
analyst): 
\begin{itemize}
\item[$-$] Which is the maximal difference $\delta_{1}$ between actions $a$
and $b$ on criterion $g_{j}$ such that $a$ and $b$ can be considered
absolutely similar with respect to the same criterion? 
\item[$-$] Which is the minimal difference $\delta_{2}$ between actions $a$
and $b$ on criterion $g_{j}$ such that $a$ and $b$ can be considered
definitely not similar with respect to the same criterion? 
\item[$-$] Which is the maximal difference $\delta_{3}$ between actions $a$
and $b$ on criterion $g_{j}$ such that there is not any dissimilarity
between $a$ and $b$ with respect to the same criterion? 
\item[$-$] Which is the minimal difference $\delta_{4}$ between actions $a$
and $b$ on criterion $g_{j}$ such that $a$ and $b$ can be considered
absolutely dissimilar with respect to the same criterion?
\end{itemize}
The above thresholds, $\delta_{1}$, $\delta_{2}$, $\delta_{3}$
and $\delta_{4}$, can be used as follows:
\begin{itemize}
\item[$-$] For values of $\Delta_{j}(a,b)$ such that $|\Delta_{j}(a,b)|\leqslant\delta_{1}$,
we have $f_{j}\big(\Delta_{j}(a,b)\big)=1$; 
\item[$-$] For values of $\Delta_{j}(a,b)$ such that $\delta_{1}<|\Delta_{j}(a,b)|\leqslant\delta_{2}$,
$f_{j}\big(\Delta_{j}(a,b)\big)$ is linear with $f_{j}\big(\Delta_{j}(a,b)\big)=1$
if $|\Delta_{j}(a,b)|=\delta_{1}$ and $f_{j}\big(\Delta_{j}(a,b)\big)=0$
if $|\Delta_{j}(a,b)|=\delta_{2}$;
\item[$-$] For values of $\Delta_{j}(a,b)$ such that $\delta_{2}<|\Delta_{j}(a,b)|\leqslant\delta_{3}$,
we have $f_{j}\big(\Delta_{j}(a,b)\big)=0$;
\item[$-$] For values of $\Delta_{j}(a,b)$ such that $\delta_{3}<|\Delta_{j}(a,b)|\leqslant\delta_{4}$,
$f_{j}\big(\Delta_{j}(a,b)\big)$ is linear with $f_{j}\big(\Delta_{j}(a,b)\big)=0$
if $|\Delta_{j}(a,b)|=\delta_{3}$ and $f_{j}\big(\Delta_{j}(a,b)\big)=-1$
if $|\Delta_{j}(a,b)|=\delta_{4}$; 
\item[$-$] For values of $\Delta_{j}(a,b)$ such that $|\Delta_{j}(a,b)|>\delta_{4}$,
we have $f_{j}\big(\Delta_{j}(a,b)\big)=-1$.
\end{itemize}

Let us observe that an alternative procedure to elicit this kind of
functions has been proposed in \citet{Costaetal2019arxiv}.

The C{\footnotesize{}AT}\nobreakdash-\hspace{0pt}SD method was designed
to take into account interaction effects between pairs of criteria
when computing likeness between two actions. In general, in real-world
problems, the following three types of interactions between criteria
can be considered \citep{figueira2009electre}: 
\begin{enumerate}
\item \emph{Mutual\nobreakdash-\hspace{0pt}strengthening effect} between
the criteria $g_{j}$ and $g_{\ell}$. This synergy effect between
the two criteria, when both criteria are in favor of similarity between
actions $a$ and $b$, can be modeled through a positive coefficient
$k_{j\ell}^{h}$ ($k_{j\ell}^{h}=k_{\ell j}^{h}$), which is added
to the sum of the weights $k_{j}^{h}+k_{\ell}^{h}$; 
\item \emph{Mutual\nobreakdash-\hspace{0pt}weakening effect} between the
criteria $g_{j}$ and $g_{\ell}$. This redundancy effect between
the two criteria, when both criteria are in favor of similarity between
actions $a$ and $b$, can be modeled through a negative coefficient
$k_{j\ell}^{h}$ ($k_{j\ell}^{h}=k_{\ell j}^{h}$), which is added
to the sum of the weights $k_{j}^{h}+k_{\ell}^{h}$; 
\item \emph{Antagonistic effect} between the criteria $g_{j}$ and $g_{\ell}$.
This antagonistic effect exercised when criterion $g_{j}$ is in favor
of the similarity and criterion $g_{p}$ is in favor of the dissimilarity
between actions $a$ and $b$, can be modeled through a negative coefficient
$k_{j|p}^{h}$, which is added to the weight $k_{j}^{h}$ (in general,
$k_{j|p}^{h}$ is not equal to $k_{p|j}^{h}$ or, even more, one of
the two antagonistic effects could not exist). 
\end{enumerate}
It should be remarked that distinct sets of weights and interaction
coefficients, $k(C_{h})$, can be defined among categories, $h=1,...,q$.
For example, let us consider a problem in which some cars have to
be assigned to categories ``family car'' and ``sport car'', and
that criteria cost, safety, maximum speed and acceleration have to
be taken into account. One can imagine that, on one hand, cost and
safety are the most important criteria when assigning a car to the
``family car'' category, while, on the other hand, maximum speed
and acceleration become the most important criteria in assigning a
car to the ``sport car'' category.

To guarantee that the contribution of each criterion to the comprehensive
similarity is not negative when considering the interaction effects,
the following net flow condition has to be fulfilled \citep{figueira2009electre}:

\begin{equation}
k_{j}^{h}\;\;\;\;-\hspace{-0.5cm}\sum_{\big\{\{j,\ell\}\in M^{h}\;:\;k_{j\ell}^{h}<0\big\}}\hspace{-1cm}\vert k_{j\ell}^{h}\vert\;\;-\sum_{(j,p)\in O^{h}}\hspace{-0.25cm}\vert k_{j|p}^{h}\vert\;\geqslant\;0{\displaystyle ,\;\mbox{for all}\;j\;\mbox{and}\;{\displaystyle h=1,...,q}},\label{eq:NetFlow}
\end{equation}
where
\begin{itemize}
\item[$-$] $M^{h}$ is the set of all pairs of criteria $\left\{ j,\ell\right\} $
such that $f_{j}\big(\Delta_{j}(a,b)\big)>0$, \textit{$f_{\ell}\big(\Delta_{\ell}(a,b)\big)>0$,
}and there is mutual-weakening effect between them, for category $C_{h}$,
$h=1,...,q$;
\item[$-$] $O^{h}$ is the set of all ordered pairs of criteria $\left(j,p\right)$
such that $f_{j}\big(\Delta_{j}(a,b)\big)>0$, \textit{$f_{p}\big(\Delta_{p}(a,b)\big)<0$,
}and $g_{j}$ exercises an antagonistic effect on $g_{p}$, for category
$C_{h}$, $h=1,...,q$.
\end{itemize}
Considering a similarity-dissimilarity function for each criterion,
the set of criteria weights and the interaction coefficients defined
for each category $C_{h}$, $h=1,...,q$, a comprehensive similarity
aggregation function can be defined. Such a function measures the
strength of the arguments in favor of likeness of action $a$ with
respect to action $b$. A \emph{comprehensive similarity function}
is a real-valued function $f^{s}\;:\;[0,1]^{n}\times[-1,0]^{n}\rightarrow[0,1]$
defined as follows:

\[
{\displaystyle s^{h}(a,b)=f^{s}\big(s_{1}(a,b),\ldots,s_{j}(a,b),\ldots,s_{n}(a,b),d_{1}(a,b),\ldots,d_{j}(a,b),\ldots,d_{n}(a,b),k(C_{h})\big)=}
\]
 
\begin{equation}
={\displaystyle \frac{1}{K^{h}(a,b)}\left(\sum_{j\in G}k_{j}^{h}s_{j}(a,b)+\sum_{\{j,\ell\}\in M^{h}}s_{j}(a,b)s_{\ell}(a,b)k_{j\ell}^{h}+\sum_{(j,p)\in O^{h}}s_{j}(a,b)|d_{p}(a,b)|k_{j|p}^{h}\right)}\label{eq:s(a,b)}
\end{equation}

\noindent and 
\[
K^{h}(a,b)=\sum_{j\in G}k_{j}^{h}+\sum_{\{j,\ell\}\in M^{h}}s_{j}(a,b)s_{\ell}(a,b)k_{j\ell}^{h}+\sum_{(j,p)\in O^{h}}s_{j}(a,b)|d_{p}(a,b)|k_{j|p}^{h},
\]

\noindent for $h=1,...,q$.

A comprehensive dissimilarity function, $d(a,b)$, can also be defined
to measure the strength of the arguments in favor of dissimilarity
between actions $a$ and $b$, i.e., in opposition to likeness. The
function considers only the dissimilarity values obtained from all
per-criterion dissimilarity functions. A \emph{comprehensive dissimilarity
function} $d(a,b)$ can be defined for each $(a,b)\in A\times A$
through a real-valued function $f^{d}:[-1,0]^{n}\rightarrow[-1,0]$
as follows:

\begin{equation}
{\displaystyle d(a,b)=f^{d}\big(d_{1}(a,b),\ldots,d_{j}(a,b),\ldots,d_{n}(a,b)\big){\displaystyle =\prod_{j=1}^{n}\big(1+d_{j}(a,b)\big)-1.}}
\end{equation}

In order to calculate a likeness degree that aggregates similarity
and dissimilarity, for each pair of actions $(a,b)$ ($a$ represents
a given action and $b$ a reference action), it is necessary to use
an aggregation function. A \emph{comprehensive likeness function}
$\delta(a,b)$ can be defined for each $(a,b)\in A\times A$ through
a real-valued function $f:[0,1]\times[-1,0]\rightarrow[0,1]$ as follows:
\begin{equation}
{\displaystyle \delta(a,b)=f\big(s^{h}(a,b),d(a,b)\big)=s^{h}(a,b)\big(1+d(a,b)\big)}.
\end{equation}

Thus, it is possible to assess the degree of likeness of action $a$
with respect to action $b$. $\delta(a,b)$ is called \textit{likeness
degree} between $a$ and $b$.


\subsection{Relation between actions and reference actions}

\noindent 
In order to assign actions to category $C_{h}$, $h=1,\ldots,q$,
each action has to be compared to each reference action, $b_{h\ell}$,
$\ell=1,...,|B_{h}$\textbar{}, computing the likeness degree, i.e.,
$\delta(a,b_{h\ell}$), between $a$ and $b_{h\ell}$. A \textit{likeness}
\textit{degree between the action $a$ and the reference set $B_{h}$}
can be defined as follows:

\noindent 
\begin{equation}
\delta(a,B_{h})=\underset{\ell=1,...,|B_{h}|}{\max}\left\{ \delta(a,b_{h\ell})\right\} .\label{eq:degree_max}
\end{equation}

A \textit{likeness threshold}, $\lambda^{h}$, can be chosen by the
DM for each category $C_{h}$, $h=1,...,q$. This preference parameter
is the minimum likeness degree considered necessary to say that an
action $a$ is similar to the set $B_{h},$ $h=1,...,q$, taking all
criteria into account. It can be interpreted as a majority measure
of likeness allowing an action to be assigned to the most adequate
categories, if any. Then, $\lambda^{h}$ takes a value within the
range $[0.5,1]$. A \textit{likeness binary relation}, $S(\lambda^{h})$,
is defined as follows:

\begin{equation}
aS(\lambda^{h})B_{h}\Leftrightarrow\delta(a,B_{h})\geqslant\lambda^{h}.\label{eq:sim_bin_relation}
\end{equation}

\subsection{Assignment procedure}

\noindent The \textsc{Cat-SD} assignment procedure provides at least
one category to which an action $a$ can be assigned. Each category
$C_{h}$, $h=1,...,q,$ is defined to receive actions to be processed
in an identical way, at least in a first step. Given $\lambda^{h}\in[0.5,1]$,
$h=1,\ldots,q$, the \textit{likeness}\emph{ assignment procedure}
was designed for \textsc{Cat-SD} as follows: 
\begin{itemize}
\item[$i)$] Compare action $a$ with set $B_{h}$, $h=1,\ldots,q$; 
\item[$ii)$] Identify $U=\{u\;:\;aS(\lambda^{u})B_{u}\}$; 
\item[$iii)$] Assign action $a$ to category $C_{u}$, for all $u\in U$; 
\item[$iv)$] If $U=\emptyset$, assign action $a$ to category $C_{q+1}$. 
\end{itemize}
The assignment of an action to a given category is independent from
the assignment to another category. Accordingly, a given action $a$
can be assigned to:
\begin{itemize}
\item[$-$] A single category (including $C_{q+1}$), in the case of $a$ being
only suitable to one category $C_{h}$, $h=1,...,q$ (or any);
\item[$-$] A set of categories (excluding $C_{q+1}$), in the case of $a$ being
suitable for more than one category.
\end{itemize}
\noindent 

\section{MCHP and the hierarchical CAT-SD method}

\noindent 
\label{MCHP_CAT_SD}
In some real-world problems, criteria are not all at the same level
but they can be structured in a hierarchical way as shown, for example,
in Fig. \ref{MCHPFig}. It is therefore possible to consider a root
criterion $g_{\mathbf{0}}$, some macro-criteria descending from the
root criterion and so on until the last level where the elementary
criteria are placed.

\noindent \begin{small} 
\begin{figure}
\centering {\includegraphics[scale=0.5]{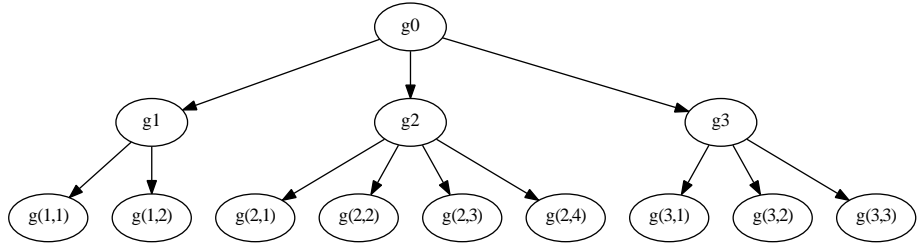}} \caption{An example of criteria structured in a hierarchical way\label{MCHPFig}}
\end{figure}

\end{small}

The MCHP has been recently introduced in literature to deal with problems
in which actions are evaluated on criteria structured in a hierarchical
way \citep{CGShierarchy,CGSoutr,CGSElectreTri}. The application of
the MCHP permits to decompose the problem in small sub-problems giving
to the DM the possibility to focus on a particular aspect of the problem
at hand. In this way, the DM can provide information at partial level,
that is considering a single criterion in the hierarchy and, at the
same time, the DM can get information on the comparisons between alternatives
taking into account the node on which he is more interested. In this
section, we shall detail the extension of the \textsc{Cat-SD} method
to the hierarchical case. Therefore, the MCHP and the \textsc{Cat-SD}
method will be put together within a unified framework giving arise
to the h\textsc{Cat-SD} method. To this aim, regarding the MCHP, we
shall use the following notation:
\begin{itemize}
\item[$-$] ${\cal G}$ is the set composed of all criteria in the hierarchy;
\item[$-$] ${\cal I}_{{\cal G}}$ is the set of the indices of criteria in ${\cal G}$;
\item[$-$] $EL\subseteq{\cal I}_{{\cal G}}$ is the set of the indices of elementary
criteria;
\item[$-$] $g_{\mathbf{r}}$, with $\mathbf{r}\in{\cal I}_{{\cal G}}\setminus EL$,
is a generic criterion in the hierarchy and it will be called \textit{non-elementary
criterion};
\item[$-$] Given a non-elementary criterion $g_{\mathbf{r}}$, $E(g_{r})\subseteq EL$
is the set of the indices of the elementary criteria descending from
$g_{\mathbf{r}}$.
\end{itemize}
Given a non-elementary criterion $g_{\mathbf{r}}$, to perform the
classification of the actions on $g_{\mathbf{r}}$, a \textit{partial
similarity function} $s_{\mathbf{r}}^{h}(a,b)$ can be defined for
each $(a,b)\in A\times A$ through $f_{\mathbf{r}}^{s}:[0,1]^{|E(g_{\mathbf{r}})|}\times[-1,0]^{|E(g_{\mathbf{r}})|}\rightarrow[0,1]$,
with $E(g_{\mathbf{r}})=\{\mathbf{t}_{1},\ldots,\mathbf{t}_{r}\}$,
as follows:

\[
s_{\mathbf{r}}^{h}(a,b)=f_{\mathbf{r}}^{s}\left(s_{\mathbf{t}_{1}}(a,b),\ldots,s_{\mathbf{t}_{r}}(a,b),d_{\mathbf{t}_{1}}(a,b),\ldots,d_{\mathbf{t}_{r}}(a,b),k(C_{h})\right)=
\]

\begin{small}  
\begin{equation}
=\frac{1}{K_{\mathbf{r}}^{h}(a,b)}\left({\displaystyle \sum_{\mathbf{t}\in E(g_{\mathbf{r}})}k_{\mathbf{t}}^{h}s_{\mathbf{t}}(a,b)+{\displaystyle \sum_{\substack{\{\mathbf{t}_{1},\mathbf{t}_{2}\}\in M^{h}:\\
\mathbf{t}_{1},\mathbf{t}_{2}\in E(g_{\mathbf{r}})
}
}s_{\mathbf{t}_{1}}(a,b)s_{\mathbf{t}_{2}}(a,b)k_{\mathbf{t}_{1}\mathbf{t}_{2}}^{h}+{\displaystyle \sum_{\substack{(\mathbf{t}_{1},\mathbf{t}_{2})\in O^{h}:\\
\mathbf{t}_{1},\mathbf{t}_{2}\in E(g_{\mathbf{r}})
}
}s_{\mathbf{t}_{1}}(a,b)|d_{\mathbf{t}_{2}}(a,b)|k_{\mathbf{t}_{1}|\mathbf{t}_{2}}^{h}}}}\right)\label{PartialSimilarityFunction}
\end{equation}
\end{small}

\noindent and  
\[
K_{\mathbf{r}}^{h}(a,b)={\displaystyle \sum_{\mathbf{t}\in E(g_{\mathbf{r}})}k_{\mathbf{t}}^{h}+{\displaystyle \sum_{\substack{\{\mathbf{t}_{1},\mathbf{t}_{2}\}\in M^{h}:\\
\mathbf{t}_{1},\mathbf{t}_{2}\in E(g_{\mathbf{r}})
}
}s_{\mathbf{t}_{1}}(a,b)s_{\mathbf{t}_{2}}(a,b)k_{\mathbf{t}_{1}\mathbf{t}_{2}}^{h}+{\displaystyle \sum_{\substack{(\mathbf{t}_{1},\mathbf{t}_{2})\in O^{h}:\\
\mathbf{t}_{1},\mathbf{t}_{2}\in E(g_{\mathbf{r}})
}
}s_{\mathbf{t}_{1}}(a,b)|d_{\mathbf{t}_{2}}(a,b)|k_{\mathbf{t}_{1}|\mathbf{t}_{2}}^{h}.}}}
\]

In this way, the partial similarity function $s_{\mathbf{r}}^{h}(a,b)$
computes the similarity between the actions $a$ and $b$ taking into
account the elementary criteria descending from $g_{\mathbf{r}}$
only.

As already done for the partial similarity function, the \textit{partial
dissimilarity function} $d_{\mathbf{r}}(a,b)$ can be defined for
each non-elementary criterion $g_{\mathbf{r}}$ and for each $(a,b)\in A\times A$
through $f_{\mathbf{r}}^{d}:[-1,0]^{|E(g_{\mathbf{r}})|}\rightarrow[-1,0]$
as follows:

\begin{equation}
d_{\mathbf{r}}(a,b)=f_{\mathbf{r}}^{d}\left(d_{\mathbf{t}_{1}}(a,b),\ldots,d_{\mathbf{t}_{r}}(a,b)\right)={\displaystyle \prod_{\mathbf{t}\in E(g_{\mathbf{r}})}\left(1+d_{\mathbf{t}}(a,b)\right)-1.}\label{PartialDissimilarityFunction}
\end{equation}

On the basis of the partial similarity and dissimilarity functions
defined in eqs. (\ref{PartialSimilarityFunction}) and (\ref{PartialDissimilarityFunction}),
for each non-elementary criterion $g_{\mathbf{r}}$ a \textit{partial
likeness function} $\delta_{\mathbf{r}}(a,b)$ can be defined for
each $(a,b)\in A\times A$ through $f_{\mathbf{r}}:[0,1]\times[-1,0]\rightarrow[0,1]$
as follows (also called likeness degree):

\begin{equation}
\delta_{\mathbf{r}}(a,b)=f_{\mathbf{r}}\left(s_{\mathbf{r}}^{h}(a,b),d_{\mathbf{r}}(a,b)\right)=s_{\mathbf{r}}^{h}(a,b)(1+d_{\mathbf{r}}(a,b)).\label{PartialCredibility}
\end{equation}

In order to assign the actions to the different categories on the
non-elementary criterion $g_{\mathbf{r}}$, these have to be compared
with the reference actions belonging to the reference set of the considered
categories. Therefore, on the basis of eq. (\ref{PartialCredibility}),
the \textit{partial likeness} \textit{degree between action $a$ and
the reference set $B_{h}$ }on $g_{\mathbf{r}}$ can be defined:

\begin{equation}
\delta_{\mathbf{r}}(a,B_{h})=\max_{l=1,\ldots,|B_{h}|}\{\delta_{\mathbf{r}}(a,b_{hl})\}.\label{SimilarityAlternativeRefSet}
\end{equation}

As a consequence, we say that $a$ is alike to $B_{h}$ on $g_{\mathbf{r}}$,
and we write $aS_{\mathbf{r}}(\lambda_{\mathbf{r}}^{h})B_{h}$, iff
$\delta_{\mathbf{r}}(a,B_{h})\geqslant\lambda_{\mathbf{r}}^{h}$,
where $\lambda_{\mathbf{r}}^{h}\in[0.5,1]$ is the likeness threshold.
Pay attention to the fact that $\lambda_{\mathbf{r}}^{h}$ can be
dependent on criterion $g_{\mathbf{r}}$ we are considering.

The \textit{partial classification} of $a\in A$ on $g_{\mathbf{r}}$
is therefore performed following these steps: 
\begin{itemize}
\item[$i)$] Compare $a$ with the set $B_{h}$ on criterion $g_{\mathbf{r}}$,
$h=1,\ldots,q,$ 
\item[$ii)$] Identify $U_{\mathbf{r}}=\{u:aS_{\mathbf{r}}(\lambda_{\mathbf{r}}^{u})B_{u}\}$, 
\item[$iii)$] Assign $a$ to the category $C_{u}$ for all $u\in U_{\mathbf{r}}$, 
\item[$iv)$] If $U_{\mathbf{r}}=\emptyset$, assign $a$ to $C_{q+1}$, being
a fictitious category collecting all non-assigned actions. 
\end{itemize}
The added value of the application of MCHP to the \textsc{Cat-SD}
method is that one can get the classifications of the actions not
only at comprehensive level, therefore considering simultaneously
all criteria, but also at a partial level by considering a particular
aspect of the problem only. In this way, the DM can have a deeper
knowledge of the decision making problem he is dealing with.


\section{The hierarchical and imprecise SRF method}

\noindent \label{ImpreciseSRF}
As described in the previous section, the classification procedure
used in the h\textsc{Cat-SD} method is based on the knowledge of the
weights of elementary criteria $g_{\mathbf{t}}$ $(k_{\mathbf{t}})$,
the knowledge of the values representing the mutual-strengthening
and mutual-weakening effects between elementary criteria $g_{\mathbf{t}_{1}},g_{\mathbf{t}_{2}}$
$\left(k_{\mathbf{t}_{1}\mathbf{t}_{2}}\right)$, and the knowledge
of the values representing the antagonistic effect exercised from
elementary criterion $g_{\mathbf{t}_{2}}$ over elementary criterion
$g_{\mathbf{t}_{1}}$ $\left(k_{\mathbf{t}_{1}|\mathbf{t}_{2}}\right)$.
Anyway, asking the DM to provide all these parameters is unreasonable
for their huge number as well as for the cognitive burden related
to the complexity of their meaning. Therefore, the application of
an indirect technique is preferable in this case.

To get the weights of criteria involved in the decision problem at
hand, in \citet{FigueiraRo02} the SRF method was proposed. The procedure,
known as cards method, extended the proposal of Simos \citep{simos_b,Simos90}
by permitting the DM to introduce the value $z$ representing the
ratio between the weight of the most important and the weight of the
least important criteria. A further extension of the SRF method was
recently introduced in \citet{Electre_CFGS}, permitting the DM to
provide imprecise information regarding both the number of cards that
should be included between two successive subsets of criteria and
the $z$-value introduced in the SRF method. The method was also applied
to hierarchical structures of criteria. In the following, we shall
briefly recall the main steps involved in the application of the SRF
method to the set $\{g_{(\mathbf{r},1)},\ldots,g_{(\mathbf{r},n(\mathbf{r}))}\}$
composed of the immediate sub-criteria of the non-elementary criterion
$g_{\mathbf{r}}$: 
\begin{enumerate}
\item Rank the criteria from the least important $L_{1}^{\mathbf{r}}$,
to the most important $L_{v}^{\mathbf{r}}$, where $v\leqslant n(\mathbf{r})$,
with the possibility of some ex-aequo; 
\item Define an interval $\left[low_{s}^{\mathbf{r}},upp_{s}^{\mathbf{r}}\right]$
in which $e_{s}^{\mathbf{r}}$ can vary, where $e_{s}^{\mathbf{r}}$
is the number of blank cards that have to be included between $L_{s}^{\mathbf{r}}$
and $L_{s+1}^{\mathbf{r}}$, with $s=1,\ldots,v-1$. The greater the
number of blank cards between $L_{s}^{\mathbf{r}}$ and $L_{s+1}^{\mathbf{r}}$,
the more important are criteria in $L_{s+1}^{\mathbf{r}}$ with respect
to criteria in $L_{s}^{\mathbf{r}}$; 
\item Define an interval $\left[z_{low}^{\mathbf{r}},z_{upp}^{\mathbf{r}}\right]$
in which $z^{\mathbf{r}}$ can vary, where $z^{\mathbf{r}}$ is the
ratio between weights of criteria in $L_{v}^{\mathbf{r}}$ and criteria
in $L_{1}^{\mathbf{r}}$. 
\end{enumerate}
Denoting by $K_{L_{s}^{\mathbf{r}}}$ the weight of a criterion in
$L_{s}^{\mathbf{r}}$, with $s=1,\ldots,v$, and by $C_{\mathbf{r}}$
the importance of a blank card introduced between two successive subsets
of criteria, the previous preference information is translated into
the following set of linear constraints (see \citealp{Electre_CFGS},
for more details):

\[
E_{\mathbf{r}}\left\{ \begin{array}{l}
\left.\begin{array}{l}
K_{L_{s+1}^{\mathbf{r}}}\geqslant K_{L_{s}^{\mathbf{r}}}+(low_{s}^{\mathbf{r}}+1)\cdot C_{\mathbf{r}},\\[1mm]
K_{L_{s+1}^{\mathbf{r}}}\leqslant K_{L_{s}^{\mathbf{r}}}+(upp_{s}^{\mathbf{r}}+1)\cdot C_{\mathbf{r}},\\[1mm]
C_{\mathbf{r}}>0,
\end{array}\right\} \mbox{for all}\;s=1,\ldots,v-1,\\
\;z_{low}^{\mathbf{r}}\cdot K_{L_{v}^{\mathbf{r}}}-K_{L_{1}^{\mathbf{r}}}\leqslant0,\\[1mm]
\;K_{L_{1}^{\mathbf{r}}}-z_{upp}^{\mathbf{r}}\cdot K_{L_{v}^{\mathbf{r}}}\leqslant0,\\[1mm]
\;K_{L_{1}^{\mathbf{r}}}>0.
\end{array}\right.
\]

Let us observe that constraints in $E_{\mathbf{r}}$ can be expressed
in terms of the weights of elementary criteria assuming that, for
each non-elementary criterion $g_{\mathbf{r}}$, $K_{\mathbf{r}}=\sum_{\mathbf{t}\in E(g_{\mathbf{r}})}k_{\mathbf{t}}$.
Moreover, for each $s=1,\ldots,v$ and for each $g_{(\mathbf{r},j)}\in L_{s}^{\mathbf{r}}$,
$K_{(\mathbf{r},j)}=K_{L_{s}^{\mathbf{r}}}$.

Concerning the parameters $k_{\mathbf{t}_{1}\mathbf{t}_{2}}$ and
$k_{\mathbf{t}_{1}|\mathbf{t}_{2}}$, with $\mathbf{t}_{1},\mathbf{t}_{2}\in EL,$
the following constraints translate the preferences of the DM:

\[
E_{int}\left\{ \begin{array}{l}
k_{\mathbf{t}_{1}\mathbf{t}_{2}}>0\;\;\mbox{if \ensuremath{g_{\mathbf{t}_{1}}} and \ensuremath{g_{\mathbf{t}_{2}}} present a mutual-strengthening effect},\\[1mm]
k_{\mathbf{t}_{1}\mathbf{t}_{2}}<0\;\;\mbox{if \ensuremath{g_{\mathbf{t}_{1}}} and \ensuremath{g_{\mathbf{t}_{2}}} present a mutual-weakening effect},\\[1mm]
k_{\mathbf{t}_{1}|\mathbf{t}_{2}}<0\;\;\mbox{if \ensuremath{g_{\mathbf{t}_{2}}} presents an antagonistic effect over \ensuremath{g_{\mathbf{t}_{1}}}}.\\[1mm]
\end{array}\right.
\]

The following technical constraints have also to be satisfied: 
\begin{description}
\item [{$(E_{Norm})$}] ${\displaystyle \sum_{\mathbf{t}\in EL}k_{\mathbf{t}}+\sum_{\{\mathbf{t}_{1},\mathbf{t}_{2}\}\subseteq EL}k_{\mathbf{t}_{1}\mathbf{t}_{2}}=100,}$ 
\item [{$(E_{Net})$}] $k_{\mathbf{t}_{1}}-{\displaystyle \sum_{\substack{\{\mathbf{t}_{1},\mathbf{t}_{2}\}\subseteq EL:\\
k_{\mathbf{t}_{1}\mathbf{t}_{2}}<0
}
}|k_{\mathbf{t}_{1}\mathbf{t}_{2}}|-{\displaystyle \sum_{\mathbf{t}_{3}\in EL}|k_{\mathbf{t}_{1}|\mathbf{t}_{3}}|\geqslant0}}$ for all $g_{\mathbf{t}_{1}}$ such that $\mathbf{t}_{1}\in EL$. 
\end{description}
Let us observe that $E_{Norm}$ is a technical constraint used only
to put an upper bound on the coefficients. This will be useful in
the sampling procedure that we will describe in the following section.
Anyway, if one uses the direct technique, that is the DM provides
directly the values of the coefficients involved in the computations,
then this constraint can be neglected. The space of the parameters
involved in the hierarchical and imprecise SRF method is therefore
defined by the constraints in the set:

\[
E={\displaystyle \cup_{\mathbf{r}\in{\cal I}_{{\cal G}}\setminus EL}E_{\mathbf{r}}\cup E_{int}\cup E_{Norm}\cup E_{Net}.}
\]

To check if there exists at least one set of parameters compatible
with the preferences provided by the DM, one has to solve the following
LP problem:

\begin{equation}
\varepsilon^{*}=\max\varepsilon,\;\mbox{subject to }E^{'}\label{LPCheck}
\end{equation}

\noindent where $E^{'}$ is obtained by $E$ converting the strict
inequality constraints in weak ones by using an auxiliary variable
$\varepsilon$. For example, constraint $C_{\mathbf{r}}>0$ is converted
into $C_{\mathbf{r}}\geqslant\varepsilon$, while $k_{\mathbf{t}_{1}\mathbf{t}_{2}}<0$
is converted into $k_{\mathbf{t}_{1}\mathbf{t}_{2}}\leqslant-\varepsilon.$
If $E^{'}$ is feasible and $\varepsilon^{*}>0$, then the space of
parameters is not empty while, in the opposite case, the set of constraints
$E^{'}$ is infeasible and the cause of the infeasibility can be checked
by using one of the methods proposed in \citet{mousseau2003resolving}.

Let us observe that the hierarchical and imprecise SRF method involves
the application of the imprecise SRF method to each node of the hierarchy.
For example, if one deals with a hierarchical structure of criteria
such that one shown in Fig. \ref{MCHPFig}, the imprecise SRF method
has to be applied at first on the set of criteria $\{g_{\mathbf{1}},g_{\mathbf{2}},g_{\mathbf{3}}\}$,
and then to the three sets of elementary criteria $\{g_{(\mathbf{1},1)},g_{(\mathbf{1},2)}\},\{g_{(\mathbf{2},1)},g_{(\mathbf{2},2)},g_{(\mathbf{2},3)},g_{(\mathbf{2},4)}\}$
and $\{g_{(\mathbf{3},1)},g_{(\mathbf{3},2)},g_{(\mathbf{3},3)}\}$.
The application of the hierarchical and imprecise SRF method will
be carefully described and illustrated in Section \ref{Example}.


\subsection{Eliciting interaction and antagonistic coefficients with SRF method}

\noindent 
In \citealp{Electre_CFGS} only the sign of the interaction coefficients
and the presence of antagonistic coefficients were considered and
coded with constraints in $E_{int}$. Instead it is possible to get
more precise preference information from the DM by considering additional
cards referred to pairs of criteria for which there is an interaction
or an antagonistic effect. More precisely: 
\begin{itemize}
\item[$-$] In case of mutual-strengthening effect between criteria $g_{i}$
and $g_{j}$, a card will be associated to the the pair of criteria
$\{g_{i},g_{j}\}$ and the value $K(\{g_{i},g_{j}\})$ assigned to
that card will represent the importance of the two criteria together
so that we have 
\[
K(\{g_{i},g_{j}\})=k_{i}+k_{j}+k_{ij}
\]
with $k_{ij}>0$ a parameters used to represent the mutual-strengthening
effect between the two criteria at hand; 
\item[$-$] In case of mutual-weakening effect between criteria $g_{i}$ and
$g_{j}$, a card will be associated to the the pair of criteria $\{g_{i},g_{j}\}$
and the value $K(\{g_{i},g_{j}\})$ assigned to that card will represent
the importance of the two criteria together so that we have 
\[
K(\{g_{i},g_{j}\})=k_{i}+k_{j}+k_{ij}
\]
with $k_{ij}>0$ a parameters used to represent the mutual-weakening
effect between the two criteria at hand; 
\item[$-$] In case of an antagonistic effect exercised by $g_{j}$ over $g_{i}$,
two cards will be associated to $g_{i}$, and they will be denoted
by $k_{i}$ and $k_{i}^{'}$. The first $(k_{i})$ denotes the importance
of $g_{i}$ when the antagonistic effect is not taken into account.
The second $(k_{i}^{'})$ denotes, instead, the importance of $g_{i}$
when $g_{j}$ exercises the antagonistic effect over it and, consequently,
\[
k_{i}^{'}=k_{i}+k_{i|j}
\]
where $k_{i|j}<0$ is a parameter representing the magnitude of the
antagonistic effects. 
\end{itemize}

In this way, applying the imprecise SRF method with the addition of
these cards, the DM can provide more precise information not only
regarding the type of interactions but also to its magnitude expressed
by the eventual presence of blank cards between successive subsets
of criteria.

In the following didactic example we shall show how the new procedure
works. Suppose that there are four criteria $g_{1},g_{2},g_{3}$ and
$g_{4}$. Assume that there is: 
\begin{itemize}
\item[$-$] A mutual-strengthening effect between $g_{3}$ and $g_{4}$; 
\item[$-$] A mutual-weakening effect between $g_{2}$ and $g_{4}$; 
\item[$-$] An antagonistic effect exercised by $g_{3}$ over $g_{4}$. 
\end{itemize}

To apply the SRF method, the DM is therefore provided with: 
\begin{itemize}
\item[$-$] A card for each criterion $g_{1},g_{2},g_{3}$ and $g_{4}$; 
\item[$-$] A card for the pairs $\{g_{3},g_{4}\}$ and $\{g_{2},g_{4}\}$ of
interacting criteria; 
\item[$-$] A card representing criterion $g_{4}$ when $g_{3}$ exercise an
antagonistic effect over it; 
\item[$-$] A certain number of blank cards that can be used to represent the
difference of importance between criteria, pairs of criteria or the
criterion $g_{4}$ subject to the antagonistic effect exercised by
$g_{3}$ over it. 
\end{itemize}

Suppose the DM provides the following order of importance with respect
to the criteria $g_{1},g_{2},g_{3}$ and $g_{4}$, the pairs of criteria
$\{g_{3},g_{4}\}$ and $\{g_{2},g_{4}\}$ and the criterion $g_{4}$
when $g_{3}$ exercises an antagonistic effect over it which is denoted
by $g'_{4}$ ($\prec$ means ``strictly more important than''):
\[
g_{3}\prec g_{1}\prec g'_{4}\prec g_{4}\prec g_{2}\prec\{g_{3},g_{4}\}\prec\{g_{2},g_{4}\}.
\]

The DM added the number of blank cards among parenthesis to increase
the difference of importance between successive subsets of criteria
or pairs of criteria: 
\[
g_{3}\;\;[1]\;\;g_{1}\;\;[2]\;\;g'_{4}\;\;[0]\;\;g_{4}\;\;[2]\;\;g_{2}\;\;[0]\;\;\{g_{3},g_{4}\}\;\;[2]\;\;\{g_{2},g_{4}\}.
\]

Let us remember that no blank cards between two consecutive criteria
or pairs of criteria does not mean that they have the same importance,
but only that their difference is minimal. The number of units between
$g_{3}$ and $\{g_{2},g_{4}\}$ is $(1+1)+(2+1)+(0+1)+(2+1)+(0+1)+(2+1)=13$.
The DM declares that the pair of criteria $\{g_{2},g_{4}\}$ is 20
times more important than $g_{3}$, that is, $z=20$, so that, giving
value 1 to $g_{3}$ and value 20 to $\{g_{2},g_{4}\}$, we get that
the value of the unit (a single card) is:

\[
u=\frac{20-1}{13}=\frac{20-1}{13}=1.4615.
\]
Consequently, considering the number of units separating two consecutive
criteria, pairs of criteria and criterion $g_{4}$ under antagonistic
effect, their importance is the following: 
\[
v(g_{3})=1,\;v(g_{1})=3.9231,\;v(g'_{4})=8.3077,\;v(g_{4})=9.7693,\;v(g_{2})=14.1539,
\]
\[
v(\{g_{3},g_{4}\}=15.6154,\;v(\{g_{2},g_{4}\}=20.
\]
Taking into account normalization $(E_{Norm})$, we get 
\[
k_{3}=3.3592,\;k_{1}=13.1783,\;k_{4}^{\prime}=27.9070,\;k_{4}=32.8165,k_{2}=47.5452,
\]
\[
K(\{g_{3},g_{4}\}=52.4548,\;K(\{g_{2},g_{4}\}=67.1835
\]
from which we get that: 
\begin{itemize}
\item[$-$] The mutual-strengthening coefficient of criteria $g_{3}$ and $g_{4}$
is 
\[
k_{34}=K(\{g_{3},g_{4}\}-k_{3}-k_{4}=16.2791,
\]
\item[$-$] The mutual-weakening coefficients of criteria $g_{2}$ and $g_{4}$
is 
\[
k_{24}=K(\{g_{2},g_{4}\}-k_{2}-k_{4}=-13.1782,
\]
\item[$-$] The antagonistic coefficient of criterion $g_{3}$ over criterion
$g_{4}$ is 
\[
k_{4|3}=k_{4}^{\prime}-k_{4}=-4.9096.
\]
\end{itemize}

Let us observe that constraints $(E_{Net})$ are satisfied, that is, 
\begin{itemize}
\item[$-$] $k_{2}+k_{24}=34.367\geqslant0$, 
\item[$-$] $k_{4}+k_{24}+k_{4|3}=29.4574\geqslant0$. 
\end{itemize}

After the positive result of this last control the weights $k_{i},i=1,2,3,4,$
the interaction coefficients $k_{24}$ and $k_{34}$, and the antagonistic
coefficient $k_{4|3}$ can be adopted and applied in a \textsc{Cat-SD}
procedure, as well as in any \textsc{Electre}, or even more in general,
outranking method considering interaction and antagonistic effect
between criteria.

In Section \ref{Example} the new proposal further extended by coupling
it with the imprecise SRF method is applied to the considered case
study.


\section{SMAA and the SMAA-hCAT-SD method}

\noindent \label{SMAASec}
As already stated in the previous section, the set of constraints
$E$ defines the space of vectors of parameters compatible with the
preferences provided by the DM. Anyway, in general, if there exists
at least one vector of parameters compatible with the preferences
of the DM, then there exists more than one. Therefore, using only
one of them could be considered arbitrary or meaningless, so that
it seems reasonable to take into consideration all compatible sets
of preference parameters. To avoid this choice, in this paper we shall
apply the SMAA (see \citealt{Lahdelma_book_greco,pelissari2019smaa},
for two surveys on SMAA; some recent extensions of the SMAA method
have been presented in \citealt{arcidiacono2018gaia,Electre_CFGS,corrente2019evaluating}).
In this section, we describe the application of SMAA to the h\textsc{Cat-SD}
method building, therefore, the SMAA-h\textsc{Cat-SD} method. It starts
from the sampling of several sets of compatible parameters. Since
the constraints in $E$ define a convex space of parameters, one can
use the Hit-And-Run (HAR) method to sample them \citep{smith1984,Tervonen2013EJOR,Tervonen2014}.
Of course, for each sampled set of parameters, a classification of
the actions at hand on the considered macro-criteria can be performed.
Denoting by ${\cal K}$ the space of the sets of parameters compatible
with the preferences provided by the DM, for each $k\in{\cal K}$,
$a\in A$, $g_{\mathbf{r}}$ and $C_{h}$, writing $a\xrightarrow[{k,\mathbf{r}}]{}C_{h}$
we mean that alternative $a$ is assigned to class $C_{h}$ on criterion
$g_{\mathbf{r}}$, considering the parameters in $k$. One can therefore
define the set ${\cal K}_{\mathbf{r}}^{h}(a)\subseteq{\cal K}$ composed
of the sets of compatible parameters for which $a$ is assigned to
$C_{h}$ with respect to $g_{\mathbf{r}}:$

\begin{equation}
{\cal K}_{\mathbf{r}}^{h}(a)=\left\{ k\in{\cal K}:a\xrightarrow[{k,\mathbf{r}}]{}C_{h}\right\} .\label{CategoryParametersAssignment}
\end{equation}

As observed in Section \ref{MCHP_CAT_SD}, each action could be assigned
to more than one category. Consequently, for each ${\cal C}\subseteq\{C_{1},\ldots,C_{q}\}$,
we can define also the following set

\begin{equation}
{\cal K}_{\mathbf{r}}^{{\cal C}}(a)=\left\{ k\in{\cal K}:\;\forall C_{h}\in{\cal C},\;a\xrightarrow[{k,\mathbf{r}}]{}C_{h}\right\} .\label{SetParametersAssignment}
\end{equation}

SMAA applied to the h\textsc{Cat-SD} method permits therefore to calculate
the approximate estimation of the probability with which an action
is assigned to a single category (or a set of categories) on criterion
$g_{\mathbf{r}}$. Formally,

\[
b_{\mathbf{r}}^{h}(a)=\frac{|{\cal K}_{\mathbf{r}}^{h}(a)|}{|{\cal K}|}\qquad\mbox{and}\qquad b_{\mathbf{r}}^{{\cal C}}(a)=\frac{|{\cal K}_{\mathbf{r}}^{{\cal C}}(a)|}{|{\cal K}|}.
\]

In this way, it is possible to analyze not only the probability of
the assignments when all elementary criteria are taken into account,
but also when a particular macro-criterion is considered.


\section{Additional requirements for the assignments\label{sec:RobustAssignment}}

\noindent The two new aspects of the approach we are proposing with
respect to the basic model presented in \citet{Costaetal2018} are
the probabilistic nature of the classification and the hierarchy of
criteria. Let us discuss their implications and their advantages.
The idea of probabilistic classification has gained a great success
in the domain of data mining and ML (see, for example, \citealt{taskar2001probabilistic,williams1998bayesian}).
The probabilistic aspect of the classification we are considering
regards the imprecision related to the weights representing the importance
of criteria, but, of course, other types of imprecision could be considered,
such as values of other parameters of the model, as the likeness thresholds
or the shape of the per-criterion similarity $s_{j}(a,b)$ through
the function $f_{j}(\Delta_{j}(a,b))$.

The robustness concerns are taken into account through a probabilistic
classification that gives, for each action, the probability to be
assigned to a given category with respect to all non-elementary criteria
in the hierarchy. However, the probability of assignment we are taking
into account is not related to the inconsistency of the elicitation.
Indeed, we have inconsistency when the information supplied by the
DM cannot be represented by the adopted decision model. This is not
the case of the probability we are using. Rather the contrary, this
probability represents the ``surplus\textquotedblright{} of possibility
to represent the information supplied by the DM for which there is
a plurality of compatible instances of the considered models. Indeed,
the probability we compute expresses the share of those instances
for which a given action is assigned to some categories with respect
to some non-elementary criteria. Therefore, SMAA has the advantage
of presenting in a clear way that on the basis of available preference
information supplied by the DM one or several classifications are
possible and, in this second case, how much one is more probable than
the others. However, in general, for fulfilling his scopes, a DM needs
one deterministic nominal classification. Consequently, there is the
need to pass from the probabilistic classification to the deterministic
classification in the most reasonable way and, in any case, taking
into account the probabilistic classification supplied by SMAA. This
is the aim of the procedure we are proposing and that provides a deterministic
nominal classification that: 
\begin{enumerate}
\item Minimizes the error of misclassification taking into account the probabilistic
information given by the application of the SMAA methodology; 
\item Fulfills some prespecified requirements related to the cardinality
of the considered categories \citep{mousseau2003notion,Kadzinski2015830,Discard,ozpeynirci2018interactive,stal2011application},
such as: 
\begin{description}
\item [{R1)}] At least $s_{h}$ alternatives should be assigned to each
category $C_{h}$, $h=1,\ldots,q$; 
\item [{R2)}] At most $s_{h}^{'}$ alternatives should be assigned to each
category $C_{h}$, $h=1,\ldots,q$; 
\item [{R3)}] At most $s_{q+1}^{'}$ alternatives should be assigned to
the category $C_{q+1}$, etc. 
\end{description}
Even if the introduced requirements could be considered ``ad hoc\char`\"{},
it is important to note that the successful application of any decision
aiding procedure depends on the appropriated customization of the
adopted formal model to the concrete decision problem at hand, so
that many important points of the formal procedure depends on the
context and must be ad hoc with respect to the specific problem. More
in general, we have to observe that the idea that all concepts in
any discipline are in some form ad hoc is gaining more and more consensus
(see, e.g., \citealt{casasanto2015all}). 
\end{enumerate}

With respect to Point 1. above, the wished deterministic nominal classification
will be obtained, for each non-elementary criterion $g_{\mathbf{r}}$,
minimizing the following loss function \citep{gneiting2007strictly,savage1971elicitation,schervish1989general}
\begin{equation}
L(\mathbf{y}_{\mathbf{r}})=\sum_{a\in A}\sum_{h=1}^{q+1}y_{a,\mathbf{r}}^{h}\sum_{k\neq h}b_{\mathbf{r}}^{k}(a)\label{LossFunction}
\end{equation}
where, for each $g_{\mathbf{r}}$, $\mathbf{r}\in{\cal I}_{{\cal G}}\setminus EL$,
$\mathbf{y}_{\mathbf{r}}=[{\ y}_{a,\mathbf{r}}^{h},\ \ a\in A,\ h=1,...,q+1]$
and ${y}_{a,\mathbf{r}}^{h}=1$ if action $a$ is assigned to category
$C_{h}$ with respect to $g_{\mathbf{r}}$, while ${y}_{a,\mathbf{r}}^{h}=0$
otherwise. 

Let us observe that, for each $a\in A$ and for each $h=1,\ldots,q+1$,
the quantity ${\displaystyle \sum_{k\neq h}b_{r}^{k}(a)}$ in eq.
(\ref{LossFunction}) represents the error made in assigning $a$
to $C_{h}$ w.r.t. $g_{\mathbf{r}}$ considering the probabilistic
information given by the SMAA methodology. For example, considering
a non-elementary criterion $g_{\mathbf{r}}$, let us assume that $a$
could be assigned to only one between $C_{1}$, $C_{2}$ and $C_{3}$
with frequencies $10\%,$ $20\%$ and $70\%$, respectively. Then,
it is obvious that the error made in assigning $a$ to the considered
categories is $90\%$ $(20\%+70\%)$, $80\%$ $(10\%+70\%)$ and $30\%$
$(10\%+20\%)$, respectively. Therefore, taking into account only
$a$ and imposing that it should be assigned to at least one category,
the minimum value of $L(\mathbf{y}_{\mathbf{r}})$ will be obtained
when $y_{a,\mathbf{r}}^{1}=y_{a,\mathbf{r}}^{2}=0$ and $y_{a,\mathbf{r}}^{3}=1.$

With respect to point 2. above, the considered requirements will be
translated into linear constraints that should be respected while
minimizing $L(\mathbf{y}_{\mathbf{r}})$. For example, assuming that
requirements \textbf{R1)}, \textbf{R2)} and \textbf{R3)} hold for
each non-elementary criterion $g_{\mathbf{r}}$, they are translated
into the constraints 
\begin{description}
\item [{C1)}] ${\displaystyle \sum_{a\in A}y_{a,\mathbf{r}}^{h}\geqslant s_{h}}$
for all $h=1,\ldots,q$; 
\item [{C2)}] ${\displaystyle \sum_{a\in A}y_{a,\mathbf{r}}^{h}\leqslant s_{h}^{'}}$
for all $h=1,\ldots,q$; 
\item [{C3)}] ${\displaystyle \sum_{a\in A}y_{a,\mathbf{r}}^{q+1}\leqslant s_{q+1}^{'}}$. 
\end{description}

To conclude this section, let us observe that more than one deterministic
nominal classification can restore the same value of the loss function
$L(\mathbf{y}_{\mathbf{r}})$. Denoting by $\mathbf{y}_{\mathbf{r}}^{*}$
the binary vector obtained as a solution of the minimization of eq.
(\ref{LossFunction}) and by $z_{\mathbf{r}}^{*}$ the number of 1s
in $\mathbf{y}_{\mathbf{r}}^{*}$, one can check for the existence
of another deterministic nominal classification respecting the provided
requirements and having the same value $L(\mathbf{y}_{\mathbf{r}}^{*})$
by minimizing eq. (\ref{LossFunction}) with subject to the constraints
translating the considered requirements with the addition of the following
ones: 
\[
\begin{array}{l}
L(\mathbf{y}_{\mathbf{r}})=L(\mathbf{y}_{\mathbf{r}}^{*}),\\[2mm]
{\displaystyle \sum_{y_{a,\mathbf{r}}^{h}\in\mathbf{y}_{\mathbf{r}}^{*}:\;y_{a,\mathbf{r}}^{h}=1}y_{a,\mathbf{r}}^{h}\leqslant z_{\mathbf{r}}^{*}-1.}
\end{array}
\]
The first constraint is used to avoid a deterioration of the optimal
value of the loss function previously found, while the second one
avoids to obtain, again, the deterministic nominal classification
previously obtained. If the LP problem is feasible, then another nominal
classification is obtained, otherwise, the previously found is unique.
By proceeding in an iterative way, it is therefore possible to obtain
all the deterministic nominal classifications minimizing the misclassification
error and respecting all the considered requirements.

\section{Illustrative example}

\noindent \label{Example}
In this section, we shall apply the SMAA-h\textsc{Cat-SD} method presented
in the previous sections extending the numerical example presented
in \citet{Costaetal2018}. In particular, the section is split in
three parts. In the first part, we shall describe, in detail, how
to perform the assignments at comprehensive level as well as on each
macro-criterion. In the second part, we shall apply the SMAA method
to the hierarchical h\textsc{Cat-SD} method commenting the obtained
results. In the third part, we shall apply the classification procedure
described in Section \ref{sec:RobustAssignment} to the numerical
example.

\subsection{Introduction of the case study and description of the computations}

\begin{figure}[!htb]
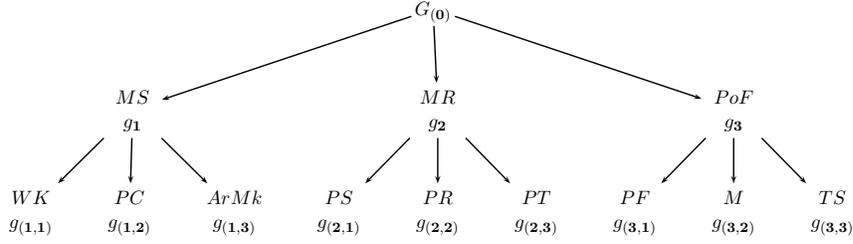

\centering \resizebox{0.7\textwidth}{!}{ \pstree[nodesep=2pt]{\TR{$G_{\mathbf{(0)}}$}}{
\pstree{\TR{$\begin{array}{c}
MS\\
g_{\mathbf{1}}
\end{array}$}}{ \TR{$\begin{array}{c}
WK\\
g_{\mathbf{(1,1)}}
\end{array}$} \TR{$\begin{array}{c}
PC\\
g_{\mathbf{(1,2)}}
\end{array}$} \TR{$\begin{array}{c}
ArMk\\
g_{\mathbf{(1,3)}}
\end{array}$} } \pstree{ \TR{$\begin{array}{c}
MR\\
g_{\mathbf{2}}
\end{array}$}}{ \TR{$\begin{array}{c}
PS\\
g_{\mathbf{(2,1)}}
\end{array}$} \TR{$\begin{array}{c}
PR\\
g_{\mathbf{(2,2)}}
\end{array}$} \TR{$\begin{array}{c}
PT\\
g_{\mathbf{(2,3)}}
\end{array}$} } \pstree{ \TR{$\begin{array}{c}
PoF\\
g_{\mathbf{3}}
\end{array}$}}{ \TR{$\begin{array}{c}
PF\\
g_{\mathbf{(3,1)}}
\end{array}$} \TR{$\begin{array}{c}
M\\
g_{\mathbf{(3,2)}}
\end{array}$} \TR{$\begin{array}{c}
TS\\
g_{\mathbf{(3,3)}}
\end{array}$} } }} 

\caption{Hierarchical structure of criteria considered in the case study\label{SoldiersExample}}
\end{figure}

\noindent Seven soldiers ($a_{1},\ldots,a_{7}$) have to be assigned
to five categories $(C_{1},\ldots,C_{5})$: snipers ($C_{1}$), breachers
($C_{2}$), communications operators ($C_{3}$), heavy weapons operators
($C_{4}$), and non-assigned candidates ($C_{5}$). Their evaluation
is performed considering several criteria structured in a hierarchical
way as shown in Fig. \ref{SoldiersExample}.

The hierarchy of criteria is composed of three macro-criteria that
are Mental Sharpness ($MS$), Mental Resilience ($MR$) and Physical
and other Features ($PoF$). Each of these macro-criteria has three
elementary criteria descending from them. In particular, World Knowledge
($WK$), Paragraph Comprehension ($PC$) and Arithmetic reasoning
and Mathematics knowledge ($ArMk$) descend from $MS$; Performance
Strategies ($PS$), Psychological Resilience ($PR$) and Personality
Traits ($PT$) descend from $MR$; finally, Physical Fitness ($PF$),
Motivation ($M$) and Teamwork Skills ($TS$) are sub-criteria of
$PoF$. The description of the nine considered elementary criteria
is given in Table \ref{ElementaryCriteriaDescription}.

\begin{table}[!htb]
\caption{Description of the elementary criteria}
\label{ElementaryCriteriaDescription}\smallskip{}
 \centering \begin{small} \resizebox{1\textwidth}{!}{ %
\begin{tabular}{lll}
\hline 
\textbf{Macro-criterion}  & \textbf{Elementary criterion}  & \textbf{Elementary criterion description} \tabularnewline
\hline 
\multirow{2}{*}{$MS$}  & $WK$  & Identification of word synonyms and right definition of words in a
given context \tabularnewline
 & $PC$  & Identification of the meaning of texts \tabularnewline
 & $ArMk$  & Solving arithmetic problems and knowledge of mathematics principles
(algebra and geometry) \tabularnewline
\hline 
\multirow{2}{*}{$MR$}  & $PS$  & Goal setting, self-talk, and emotional control \tabularnewline
 & $PR$  & Acceptance of life situations, and ability for dealing with cognitive
challenges and threats \tabularnewline
 & $PT$  & Character traits such as adaptability, dutifulness, social orientation,
self-reliance, stress tolerance, \tabularnewline
 &  & vigilance, and impulsivity \tabularnewline
\hline 
\multirow{2}{*}{$PoF$ }  & $PF$  & Physical ability with respect to aerobic fitness and strength \tabularnewline
 & $M$  & Self motivation, persistence, and dedication \tabularnewline
 & $TS$  & Communication skills and camaraderie \tabularnewline
\hline 
\end{tabular}} \end{small} 
\end{table}

The performance of the seven soldiers on the nine elementary criteria
is given in Table \ref{PerformanceSoldiers}.

\begin{table}[!htb]
\caption{Performance of the considered soldiers on the elementary criteria
at hand}
\label{PerformanceSoldiers}\smallskip{}
 \centering \begin{small} \resizebox{0,75\textwidth}{!}{ %
\begin{tabular}{llllllllll}
\hline 
\textbf{Soldier}  & $g_{(\mathbf{1},1)}$  & $g_{(\mathbf{1},2)}$  & $g_{(\mathbf{1},3)}$  & $g_{(\mathbf{2},1)}$  & $g_{(\mathbf{2},2)}$  & $g_{(\mathbf{2},3)}$  & $g_{(\mathbf{3},1)}$  & $g_{(\mathbf{3},2)}$  & $g_{(\mathbf{3},3)}$ \tabularnewline
\hline 
$a_{1}$  & 75  & 75  & 90  & 3  & 4  & 4  & 740  & 6  & 4 \tabularnewline
$a_{2}$  & 67  & 80  & 73  & 3  & 3  & 3  & 760  & 5  & 6 \tabularnewline
$a_{3}$  & 60  & 70  & 70  & 4  & 3  & 3  & 770  & 5  & 6 \tabularnewline
$a_{4}$  & 80  & 90  & 75  & 2  & 3  & 3  & 880  & 4  & 5 \tabularnewline
$a_{5}$  & 65  & 65  & 70  & 3  & 2  & 3  & 870  & 6  & 6 \tabularnewline
$a_{6}$  & 70  & 75  & 85  & 4  & 3  & 4  & 750  & 5  & 4 \tabularnewline
$a_{7}$  & 75  & 70  & 70  & 4  & 3  & 3  & 710  & 5  & 6 \tabularnewline
\hline 
\textbf{Function}  & $f_{2}$  & $f_{2}$  & $f_{2}$  & $f_{3}$  & $f_{3}$  & $f_{3}$  & $f_{1}$  & $f_{3}$  & $f_{3}$ \tabularnewline
\hline 
\end{tabular}} \end{small} 
\end{table}

Each reference set $B_{h}$ is composed of one reference action only.
Their evaluations are provided in Table \ref{PerformanceReferenceActions}.

\begin{table}[!htb]
\caption{Performance of the reference soldiers on the elementary criteria at
hand}
\label{PerformanceReferenceActions}\smallskip{}
 \centering \begin{small} \resizebox{1\textwidth}{!}{ %
\begin{tabular}{lllllllllll}
\hline 
\textbf{Reference set}  & \textbf{Reference action}  & $g_{(\mathbf{1},1)}$  & $g_{(\mathbf{1},2)}$  & $g_{(\mathbf{1},3)}$  & $g_{(\mathbf{2},1)}$  & $g_{(\mathbf{2},2)}$  & $g_{(\mathbf{2},3)}$  & $g_{(\mathbf{3},1)}$  & $g_{(\mathbf{3},2)}$  & $g_{(\mathbf{3},3)}$ \tabularnewline
\hline 
$B_{1}$  & $b_{11}$  & 80  & 75  & 85  & 4  & 4  & 4  & 700  & 6  & 4 \tabularnewline
$B_{2}$  & $b_{21}$  & 70  & 70  & 75  & 3  & 3  & 3  & 800  & 6  & 6 \tabularnewline
$B_{3}$  & $b_{31}$  & 80  & 90  & 85  & 2  & 2  & 3  & 950  & 4  & 4 \tabularnewline
$B_{4}$  & $b_{41}$  & 60  & 65  & 65  & 3  & 3  & 3  & 700  & 5  & 6 \tabularnewline
\hline 
\end{tabular}} \end{small} 
\end{table}

\begin{figure}
\centering {\includegraphics[scale=0.8]{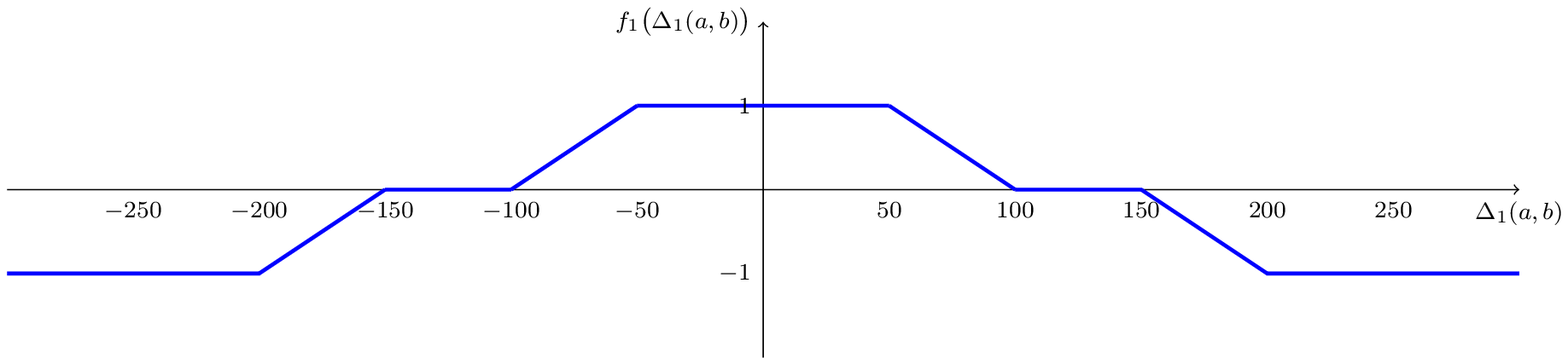}} \caption{Per-criterion similarity-dissimilarity function $f_{1}$\label{Fig:f1}}
\end{figure}

\begin{figure}
\centering {\includegraphics[scale=0.8]{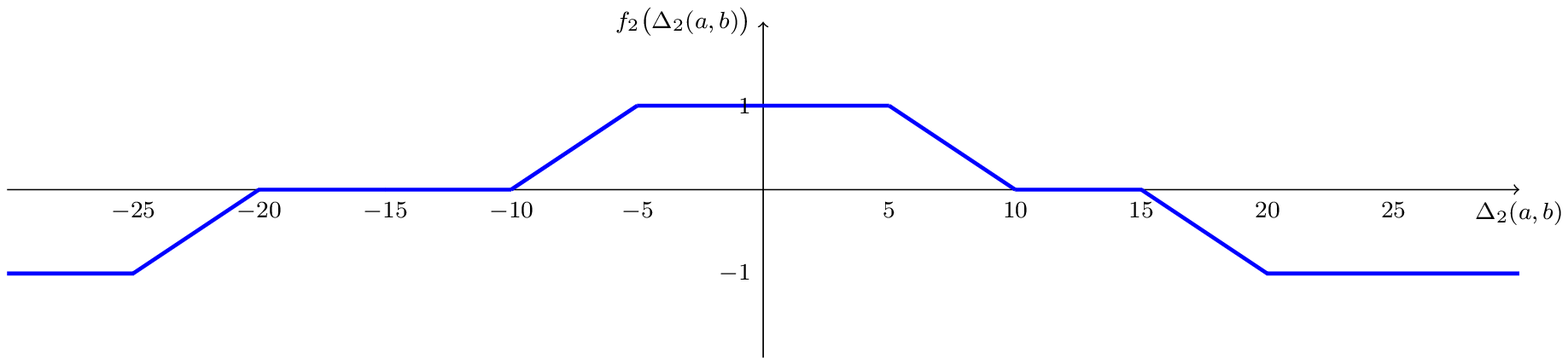}} \caption{Per-criterion similarity-dissimilarity function $f_{2}$\label{Fig:f2}}
\end{figure}

\begin{figure}
\centering {\includegraphics[scale=0.8]{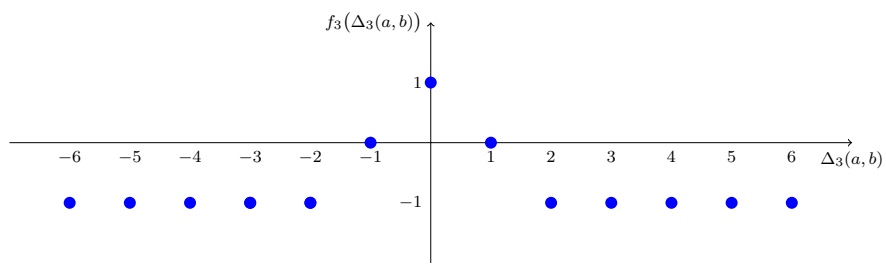}} \caption{Per-criterion similarity-dissimilarity function $f_{3}$\label{Fig:f3}}
\end{figure}

The three per-criterion similarity-dissimilarity functions used in
the illustrative example are the following (see also the graphical
representation of the functions in Figures \ref{Fig:f1}-\ref{Fig:f3}):

\begin{eqnarray*}
f_{1}\big(\Delta_{1}(a,b)\big) & = & \begin{cases}
1, & \mbox{{if}}\mbox{ }|\Delta_{1}(a,b))|\leqslant50;\\
\\
\frac{100-|\Delta_{1}(a,b)|}{50}, & \mbox{\mbox{{if}} }50<|\Delta_{1}(a,b)|\leqslant100;\\
\\
0, & \mbox{{if}}\mbox{ }100<|\Delta_{1}(a,b)|\leqslant150;\\
\\
\frac{150-|\Delta_{1}(a,b)|}{50}, & \mbox{\mbox{{if}}\mbox{ }}150<|\Delta_{1}(a,b)|\leqslant200;\\
\\
-1, & \mbox{{if}}\mbox{ }|\Delta_{1}(a,b)|>200.
\end{cases}
\end{eqnarray*}
\vspace{0.25cm}
 
\[
f_{2}\big(\Delta_{2}(a,b)\big)=\begin{cases}
1, & \mbox{\mbox{{if}}\mbox{ }}|\Delta_{2}(a,b)|\leqslant5;\\
\\
\frac{10-|\Delta_{2}(a,b)|}{5}, & \mbox{{if}}\mbox{ }5<|\Delta_{2}(a,b)|\leqslant10;\\
\\
0, & \mbox{{if}}\mbox{ }-20<\Delta_{2}(a,b))<-10\:\:\mbox{{or}}\:\:10<\Delta_{2}(a,b)\leqslant15;\\
\\
\frac{20+\Delta_{2}(a,b)}{5}, & \mbox{\mbox{{if}}\mbox{ }}-25<\Delta_{2}(a,b)\leqslant-20;\\
\\
\frac{15-\Delta_{2}(a,b)}{5}, & \mbox{\mbox{{if}}\mbox{ }}15<\Delta_{2}(a,b)\leqslant20;\\
\\
-1, & \mbox{\mbox{{if}}\mbox{ } }\Delta_{2}(a,b)\leqslant-25\:\mbox{{or}}\:\Delta_{2}(a,b)>20.
\end{cases}
\]
\vspace{0.25cm}

\[
f_{3}\big(\Delta_{3}(a,b)\big)=\begin{cases}
1, & \mbox{{if}}\mbox{ }|\Delta_{3}(a,b)|=0;\\
\\
0, & \mbox{\mbox{{if}}\mbox{ }}|\Delta_{3}(a,b)|=1;\\
\\
-1, & \mbox{\mbox{{if}}\mbox{ }}|\Delta_{3}(a,b)|\geqslant2.
\end{cases}
\]
\vspace{0.25cm}

To get the weights of the elementary criteria and the interaction
coefficients, the hierarchical and imprecise SRF method has been applied
for each category. In particular, the imprecise SRF method is applied
to the set composed by the three macro-criteria as well as to three
subsets of the elementary criteria descending from each macro-criterion.
Anyway, since the DM provided some information regarding interactions
and antagonistic effects between few elementary criteria, we had to
adapt the imprecise and hierarchical SRF method as we shall describe
in a detailed way in the following. 

Suppose that the DM provided the following information:  
\begin{enumerate}
\item There is a mutual-strengthening effect between $ArMk$ and $PR$; 
\item There is a mutual-weakening effect between $PF$ and $TS$; 
\item There is an antagonistic effect exercised by $PF$ over $PS$. 
\end{enumerate}
\noindent Each of the previous three pieces of preference information
implies a small modification in the application of the imprecise SRF
method:
\begin{itemize}
\item[$-$] In consequence of the first piece of preference information, a mutual-strengthening
effect between $MS$ and $MR$ exists too. Therefore, in applying
the SRF method at the first level, that is the level composed of criteria
$\{MS,MR,PoF\}$, the DM is provided with an additional card with
the name of the two criteria $MS$ and $MR$ on, to consider their
importance together. Then, the SRF method will be applied to the set
composed now of 4 cards $\{MS,MR,\{MS,MR\},PoF\}$. From a technical
point of view, in addition to the weights $K_{MS},$ $K_{MR},$ and
$K_{PoF}$ representing the importance of criteria $MS$, $MR$ and
$PoF$, respectively, we shall take into account also the weight $K({\{MS,MR\}})$.
In consequence of the mutual-strengthening effect between $ArMk$
and $PR$, we have that 
\[
K({\{MS,MR\}})=K_{MS}+K_{MR}+k_{ArMk,PR},
\]
where $k_{ArMk,PR}>0$ represents, indeed, the value of this effect.
Of course, $K({\{MS,MR\}})>K_{MS}$ and $K({\{MS,MR\}})>K_{MR}$;
\item[$-$] Since elementary criteria $PF$ and $TS$ descend from the same macro-criterion
$PoF$, the mutual-weakening effect between them is considered adding
another card for the pair $\{PF,TS\}$ to take into account their
importance together. The imprecise SRF method will be therefore applied
to the set $\{PF,M,TS,\{PF,TS\}\}$. The weight of the pair of criteria
$\{PF,TS\}$, that is $K(\{PF,TS\})$, will be such that 
\[
K(\{PF,TS\})=k_{PF}+k_{TS}+k_{PF,TS}
\]
where, $k_{PF,TS}<0$ is a parameter representing the mutual-weakening
effect between them; of course, in consequence of the net flow condition
(\ref{eq:NetFlow}), $K({PF,TS})>k_{PF}$ and $K({PF,TS})>k_{TS}$; 
\item[$-$] Finally, in consequence of the antagonistic effect exercised by $PF$
over $PS$, the original weight of $PS$ will be reduced. The DM is
therefore asked to apply the SRF method to the subset of criteria
$\{PS,PS^{'},PR,PT\}$, where $K(PS^{'})$ is the importance of criterion
$PS$ when $PF$ is exercising its antagonistic effect over it. Consequently,
we have 
\[
K(PS^{'})=k_{PS}+k_{PS|PF}
\]
where $k_{PS|PF}<0$ represents the magnitude of the antagonistic
effect. In this way, if the DM, for example, in applying the SRF method
will order $PS^{'}$ after $PR$, then this means that $PS$ is more
important than $PR$ even if there is another criterion $(PF)$ opposing
to it. 
\end{itemize}
In the following we shall describe in detail the application of the
hierarchical and imprecise SRF method to the sets of criteria $\{MS,MR,\{MS,MR\},PoF\}$,
$\{PF,M,TS,\{PF,TS\}\}$, $\{PS,PS^{'},PR,PT\}$ and $\{WK,PC,ArMK\}$
for each of the four categories (see also Table \ref{Tab:Imprecise_SRF}
summarizing this information). 
\begin{enumerate}
\item Application of the imprecise SRF method for category $C_{1}$:
\begin{itemize}
\item[$-$] $PoF$ is less important than $MR$ that is less important than $MS$
that, in turn, is less important than $\{MS,MR\}$. The number of
blank cards to be inserted between $\{MS,MR\}$ and $MS$ belongs
to the interval $[2,3]$; the number of blank cards between $MS$
and $MR$ varies in the interval $[1,2]$, while there is one blank
card between $MR$ and $PoF$. The ratio between the weight of $\{MS,MR\}$
and the weight of $PoF$ belongs to the interval $[4,6]$;
\item[$-$] With respect to macro-criterion $MS$, $WK$ is less important than
$PC$ that, in turn, is less important than $ArMk$. The number of
blank cards inserted between $ArMk$ and $PC$ belongs to the interval
$[0,2]$, while the number of blank cards inserted between $PC$ and
$WK$ has to belong to the interval $[0,1]$. Moreover the ratio between
the weight of $ArMk$ and the weight of $WK$ is $3$;
\item[$-$] Considering macro-criterion $MR$, $PS^{'}$ is less important than
$PS$; $PS$ is less important than $PR$ that, in turn, is less important
than $PT$. There is one blank card between $PT$ and $PR$, while
the number of blank cards to be inserted between $PR$ and $PS$ has
to belong to the interval $[1,2]$. The number of blank cards to be
included between $PS$ and $PS^{'}$ has to belong to the interval
$[0,1]$. Finally, the ratio between the weight of $PT$ and that
one of $PS^{'}$ has to belong to the interval $[2,3]$;
\item[$-$] On macro-criterion $PoF$, $M$ is less important than $PF$ being
less important than $TS$ that, in turn, is less important than $\{PF,TS\}$.
The number of blank cards between criteria in consecutive ranks varies
always in the interval $[1,2]$. The ratio between the weight of the
most important criterion ($\{PF,TS\}$) and the least important one
$(M)$ varies in the interval $[4,6]$.
\end{itemize}
\item Application of the imprecise SRF method for category $C_{2}$:
\begin{itemize}
\item[$-$] $PoF$ is less important than $MS$ that is less important than $MR$
that, in turn, is less important than $\{MR,MS\}$. The number of
blank cards to be inserted between $\{MR,MS\}$ and $MR$ belongs
to the interval $[2,3]$; there is one blank card between $MR$ and
$MS$, while the number of blank cards to be inserted between $MS$
and $PoF$ is in the interval $[1,2]$. Moreover, the ratio between
the weight of $\{MR,MS\}$ and the weight of $PoF$ is 6;
\item[$-$] With respect to macro-criterion $MS$, $WK$ is less important than
$PC$ that, in turn, is less important than $ArMk$. There is not
any blank card between $ArMk$ and $PC$. The number of blank cards
between $PC$ and $WK$ belongs to the interval $[1,2]$. Finally,
the ratio between the weight of $ArMk$ and the weight of $WK$ is
in the interval $[3,5]$;
\item[$-$] Considering macro-criterion $MR$, $PS^{'}$ is less important than
$PS$ that is less important than $PR$ that, in turn, is less important
than $PT$. The number of blank cards between $PT$ and $PR$ belongs
to the interval $[0,1]$. There is not any blank card between $PR$
and $PS$, while the number of blank cards to be inserted between
$PS$ and $PS^{'}$ varies in the interval $[0,1]$. Finally, the
ratio between the weight of $PT$ and that one of $PS^{'}$ has to
belong to the interval $[3,4]$;
\item[$-$] On macro-criterion $PoF$, $PF$ is less important than $M$ that
is less important than $TS$ being, in turn, less important than $\{PF,TS\}$.
There is one blank card between $\{PF,TS\}$ and $TS$; the number
of blank cards between $TS$ and $M$ belongs to the interval $[2,3]$,
while the number of blank cards between $M$ and $PF$ varies in the
interval $[1,2]$. Finally, the ratio between the weight of the most
important criterion ($\{PF,TS\}$) and the least important one $(PF)$
is in the interval $[3,6]$.
\end{itemize}
\item Application of the imprecise SRF method for category $C_{3}$:
\begin{itemize}
\item[$-$] $MS$ is less important than $MR$ that is less important than $\{MS,MR\}$
that, in turn, is less important than $PoF$. There is one blank card
between $PoF$ and $\{MS,MR\}$. The number of blank cards to be inserted
between $\{MS,MR\}$ and $MR$ as well as between $MR$ and $MS$
belongs to the interval $[2,3]$. Finally, the ratio between the weight
of $PoF$ and the weight of $MS$ belongs to the interval $[4,6]$;
\item[$-$] With respect to macro-criterion $MS$, $PC$ is less important than
$WK$ that is as important as $ArMk$. There is only one blank card
between $PC$ and $ArMk$. Moreover the ratio between the weight of
$ArMk$ and the weight of $PC$ is in the interval $[2,4]$;
\item[$-$] Considering macro-criterion $MR$, $PS^{'}$ is less important than
$PR$ that is less important than $PS$ having the same importance
of $PT$. There is one blank card between $PS$ and $PR$, while the
number of blank cards between $PR$ and $PS'$ belongs to the interval
$\left[0,1\right]$. Finally, the ratio between the weight of $PT$
and that one of $PS^{'}$ has to belong to the interval $[2,4]$;
\item[$-$] On macro-criterion $PoF$, $TS$ is less important than $PF$ having
the same importance of $M$ that, in turn, is less important than
the criteria in $\{PF,TS\}$. The number of blank cards between $\{PF,TS\}$
and $PF$ belongs to the interval $[0,1]$. The number of blank cards
between $PF$ and $TS$ should vary in the interval $[1,2]$. Finally,
the ratio between the weight of $\{PF,TS\}$ and the weight of $TS$
should be in the interval $[3,4]$.
\end{itemize}
\item Application of the imprecise SRF method for category $C_{4}$:
\begin{itemize}
\item[$-$] $PoF$ is less important than $MS$ that is less important than $MR$
that, in turn, is less important than $\{MS,MR\}$. The number of
blank cards between $\{MS,MR\}$ and $MR$ belongs to the interval
$[1,2]$. There is one blank card between $MR$ and $MS$. Moreover,
the number of blank cards between $MS$ and $PoF$ should vary in
the interval $[1,2]$. Finally, the ratio between the weight of $\{MS,MR\}$
and the weight of $PoF$ is 9;
\item[$-$] With respect to macro-criterion $MS$, $PC$ and $WK$ are equally
important but they are less important than $ArMk$. The number of
blank cards that should be inserted between $ArMk$ and the set of
criteria $\{PC,WK\}$ belongs to the interval $[1,2]$. Finally, the
ratio between the weight of $ArMk$ and the weight of $PC$ is 4;
\item[$-$] Considering macro-criterion $MR$, $PS^{'}$ is less important than
$PS$ that is less important than $PR$ being less important than
$PT$. The number of blank cards between $PT$ and $PR$ belongs to
the interval $[0,1]$. There is not any blank card between $PR$ and
$PS$ while the number of blank cards between $PS$ and $PS'$ belongs
to the interval $[0,1]$. Finally, the ratio between the weight of
$PT$ and that one of $PS^{'}$ has to belong to the interval $[2,4]$;
\item[$-$] On macro-criterion $PoF$, $PF$ is less important than $M$ having
the same importance of $TS$ that, in turn, is less important than
$\{PF,TS\}$. The number of blank cards between $\{PF,TS\}$ and $M$
belongs to the interval $[1,3]$; the number of blank cards between
$M$ and $PF$ varies in the interval $[1,2]$. Finally, the ratio
between the weight of $\{PF,TS\}$ and the weight of $PF$ is in the
interval $[3,5]$. 
\end{itemize}
\end{enumerate}
\begin{table}[!htb]
\caption{Data used in the hierarchical and imprecise SRF }
\label{Tab:Imprecise_SRF}\smallskip{}
 \centering \begin{small} \resizebox{1\textwidth}{!}{ %
\begin{tabular}{c|ccc|ccc|ccc|ccc}
\hline 
 & \multicolumn{3}{c|}{$C_{1}$} & \multicolumn{3}{c|}{$C_{2}$} & \multicolumn{3}{c|}{$C_{3}$} & \multicolumn{3}{c}{$C_{4}$}\tabularnewline
\hline 
Rank  & Criterion  & No. blank cards  & $z$  & Criterion  & No. blank cards  & $z$  & Criterion  & No. blank cards  & $z$  & Criterion  & No. blank cards  & $z$\tabularnewline
\hline 
1  & $\{MR,MS\}$  & $[2,3]$  & $[4,6]$  & $\{MS,MR\}$  & $[2,3]$  & $6$  & $PoF$  & $1$  & $[4,6]$  & $\{MS,MR\}$  & $[1,2]$  & $9$\tabularnewline
2  & $MS$  & $[1,2]$  &  & $MR$  & $1$  &  & $\{MS,MR\}$  & $[2,3]$  &  & $MR$  & $1$  & \tabularnewline
3  & $MR$  & $1$  &  & $MS$  & $[1,2]$  &  & $MR$  & $[2,3]$  &  & $MS$  & $[1,2]$  & \tabularnewline
4  & $PoF$  &  &  & $PoF$  &  &  & $MS$  &  &  & $PoF$  &  & \tabularnewline
\hline 
1  & $ArMk$  & $[0,2]$  & $3$  & $ArMk$  & $0$  & $[3,5]$  & $WK$, $ArMk$  & $1$  & $[2,4]$  & $ArMk$  & $[1,2]$  & $4$\tabularnewline
2  & $PC$  & $[0,1]$  &  & $PC$  & $[1,2]$  &  & $PC$  &  &  & $PC$, $WK$  &  & \tabularnewline
3  & $WK$  &  &  & $WK$  &  &  &  &  &  &  &  & \tabularnewline
\hline 
1  & $PT$  & $1$  & $[2,3]$  & $PT$  & $[0,1]$  & $[3,4]$  & $PS$, $PT$  & $1$  & $[2,4]$  & $PT$  & $[0,1]$  & $[2,4]$\tabularnewline
2  & $PR$  & $[1,2]$  &  & $PR$  & $0$  &  & $PR$  & $[0,1]$  &  & $PR$  & $0$  & \tabularnewline
3  & $PS$  & $[0,1]$  &  & $PS$  & $[0,1]$  &  & $PS^{'}$  &  &  & $PS$  & $[0,1]$  & \tabularnewline
4  & $PS^{'}$  &  &  & $PS^{'}$  &  &  &  &  &  & $PS^{'}$  &  & \tabularnewline
\hline 
1  & $\{PF,TS\}$  & $[1,2]$  & $[4,6]$  & $\{PF,TS\}$  & $1$  & $[3,6]$  & $\{PF,TS\}$  & $[0,1]$  & $[3,4]$  & $\{PF,TS\}$  & $[1,3]$  & $[3,5]$\tabularnewline
2  & $TS$  & $[1,2]$  &  & $TS$  & $[2,3]$  &  & $PF$, $M$  & $[1,2]$  &  & $M,$ $TS$  & $[1,2]$  & \tabularnewline
3  & $PF$  & $[1,2]$  &  & $M$  & $[1,2]$  &  & $TS$  &  &  & $PF$  &  & \tabularnewline
3  & $M$  &  &  & $PF$  &  &  &  &  &  &  &  & \tabularnewline
\hline 
\end{tabular}} \end{small} 
\end{table}

Introducing all the constraints translating the preference information
provided by the DM, we solved the LP problem (\ref{LPCheck}) obtaining
$\varepsilon^{*}>0.$ Therefore, there exists at least one set of
parameters compatible with the preferences provided by the DM and,
consequently, we applied the HAR method to sample 100,000 sets of
compatible parameters for each of the four categories.

Now, we shall present in detail all the steps necessary to perform
the considered assignments, highlighting the meaning of using the
MCHP. For this reason, we consider the soldier $a_{3}$ and the set
of sampled weights in Table \ref{WeightsExample}

\begin{table}[!htb]
\caption{Weights considered in the first part of the example}
\label{WeightsExample}\smallskip{}
 \centering \begin{small} \resizebox{0,9\textwidth}{!}{ %
\begin{tabular}{lccccccccc}
\hline 
 & $g_{(1,1)}$  & $g_{(1,2)}$  & $g_{(1,3)}$  & $g_{(2,1)}$  & $g_{(2,2)}$  & $g_{(2,3)}$  & $g_{(3,1)}$  & $g_{(3,2)}$  & $g_{(3,3)}$ \tabularnewline
\hline 
$k_{\mathbf{t}}^{1}$  & 8.925  & 16.269  & 26.777  & 7.537  & 11.537  & 14.347  & 5.312  & 2.361  & 8.133 \tabularnewline
$k_{\mathbf{t}}^{2}$  & 4.809  & 12.621  & 16.140  & 13.301  & 16.621  & 22.599  & 2.615  & 4.508  & 7.274 \tabularnewline
$k_{\mathbf{t}}^{3}$  & 5.033  & 2.304  & 5.033  & 12.239  & 7.082  & 12.239  & 23.561  & 23.561  & 9.320 \tabularnewline
$k_{\mathbf{t}}^{4}$  & 5.557  & 5.557  & 22.231  & 15.011  & 18.649  & 22.708  & 1.838  & 4.083  & 4.083 \tabularnewline
\hline 
\end{tabular}} \end{small} 
\end{table}

The steps that have to be performed in the assignment procedure are
the following: 
\begin{enumerate}
\item \textit{Compute the similarity-dissimilarity:} For each elementary
criterion and using the three per-criterion similarity-dissimilarity
functions introduced above, we compute the similarity-dissimilarity
between $a_{3}$ and the four reference actions. The values are shown
in Table \ref{SDCriterion}.

\begin{table}[!htb]
\caption{Similarity-dissimilarity values for each elementary criterion}
\label{SDCriterion}\smallskip{}
 \centering \begin{small} \resizebox{0,8\textwidth}{!}{ %
\begin{tabular}{llllllllll}
\hline 
 & $g_{(1,1)}$  & $g_{(1,2)}$  & $g_{(1,3)}$  & $g_{(2,1)}$  & $g_{(2,2)}$  & $g_{(2,3)}$  & $g_{(3,1)}$  & $g_{(3,2)}$  & $g_{(3,3)}$ \tabularnewline
\hline 
$f_{\mathbf{t}}(a_{3},b_{11})$  & 0  & 1  & 0  & 1  & 0  & 0  & 0.6  & 0  & -1 \tabularnewline
$f_{\mathbf{t}}(a_{3},b_{21})$  & 0  & 1  & 1  & 0  & 1  & 1  & 1  & 0  & 1 \tabularnewline
$f_{\mathbf{t}}(a_{3},b_{31})$  & 0  & 0  & 0  & -1  & 0  & 1  & -0.6  & 0  & -1 \tabularnewline
$f_{\mathbf{t}}(a_{3},b_{41})$  & 1  & 1  & 1  & 0  & 1  & 1  & 0.6  & 1  & 1 \tabularnewline
\hline 
\end{tabular}} \end{small} 
\end{table}
\item \textit{Compute the comprehensive likeness:} Following eqs. (\ref{PartialSimilarityFunction})-(\ref{PartialCredibility}),
for each non-elementary criterion $g_{\mathbf{r}}$ in the hierarchy,
we compute the partial similarity and dissimilarity functions as well
as the partial likeness degree between $a_{3}$ and the considered
reference actions. The values are shown in Table \ref{PartSD}.

\begin{table}[!htb]
\caption{Partial similarity, dissimilarity, and likeness degree}
\label{PartSD}\smallskip{}
 \centering\begin{small}\resizebox{1\textwidth}{!}{ %
\begin{tabular}{l|lll|lll|lll|lll}
\hline 
 & $s_{\mathbf{1}}^{h}(a_{3},\cdot)$  & $d_{\mathbf{1}}(a_{3},\cdot)$  & $\delta_{\mathbf{1}}(a_{3},\cdot)$  & $s_{\mathbf{2}}^{h}(a_{3},\cdot)$  & $d_{\mathbf{2}}(a_{3},\cdot)$  & $\delta_{\mathbf{2}}(a_{3},\cdot)$  & $s_{\mathbf{3}}^{h}(a_{3},\cdot)$  & $d_{\mathbf{3}}(a_{3},\cdot)$  & $\delta_{\mathbf{3}}(a_{3},\cdot)$  & $s_{\mathbf{0}}^{h}(a_{3},\cdot)$  & $d_{\mathbf{0}}(a_{3},\cdot)$  & $\delta_{\mathbf{0}}(a_{3},\cdot)$\tabularnewline
\hline 
$b_{11}$  & 0.313  & 0  & 0.313  & 0.225  & 0  & 0.225  & 0.201  & -1  & 0  & 0.266  & -1  & 0 \tabularnewline
$b_{21}$  & 0.856  & 0  & 0.856  & 0.746  & 0  & 0.543  & 0.668  & 0  & 0.668  & 0.773  & 0  & 0.773 \tabularnewline
$b_{31}$  & 0  & 0  & 0  & 0.387  & -1  & 0  & 0  & -1  & 0  & 0  & -1  & 0 \tabularnewline
$b_{41}$  & 1  & 0  & 1  & 0.733  & 0  & 0.733  & 0.926  & 0  & 0.926  & 0.842  & 0  & 0.842 \tabularnewline
\hline 
\end{tabular}} \end{small} 
\end{table}

For example, to compute $s_{\mathbf{3}}^{h}(a_{3},b_{11})$ w.r.t.
category $C_{1}$, that is the similarity between $a_{3}$ and $b_{11}$
on $PoF$ ($g_{\mathbf{3}}$) for assigning $a_{3}$ to snipers, we
have to take into account only the last three elementary criteria
as well as the mutual-strengthening effect between $PF$ ($g_{(\mathbf{3},1)}$)
and $TS$ ($g_{(\mathbf{3},3)}$). In particular, observing that $d_{\mathbf{t}}(a_{3},b_{11})=f_{\mathbf{t}}(a_{3},b_{11})$
if $f_{\mathbf{t}}(a_{3},b_{11})<0$ and 0 otherwise, and that $s_{\mathbf{t}}(a_{3},b_{11})=f_{\mathbf{t}}(a_{3},b_{11})$
if $f_{\mathbf{t}}(a_{3},b_{11})>0$ and 0 otherwise, we have that
$s_{(\mathbf{3},1)}(a_{3},b_{11})=f_{(\mathbf{3},1)}(a_{3},b_{11})$
and $d_{(\mathbf{3},3)}(a_{3},b_{11})=f_{(\mathbf{3},3)}(a_{3},b_{11})$.
We obtain:
\begin{itemize}
\item[$-$] $K_{\mathbf{3}}^{{h}}(a,b_{11})=k_{(\mathbf{3},1)}^{1}+k_{(\mathbf{3},2)}^{1}+k_{(\mathbf{3},3)}^{1}=5.312+2.361+8.133=15.806;$
\item[$-$] $s_{\mathbf{3}}^{h}(a_{3},b_{11})=\frac{k_{(\mathbf{3},1)}^{{1}}\cdot f_{(\mathbf{3},1)}(a_{3},b_{11})}{K_{\mathbf{3}}^{h}(a,b_{11})}=\frac{5.312\cdot0.6}{15.806}=0.2016;$
\item[$-$] $d_{\mathbf{3}}(a_{3},b_{11})=\left(1+d_{(\mathbf{3},3)}(a_{3},b_{11})\right)-1=d_{(\mathbf{3},3)}(a_{3},b_{11})=-1$;
\item[$-$] $\delta_{\mathbf{3}}(a_{3},b_{11})=s_{\mathbf{3}}^{h}(a_{3},b_{11})\left(1+d_{\mathbf{3}}(a_{3},b_{11})\right)=0.2016\cdot(1-1)=0.$
\end{itemize}
The other values in Table \ref{PartSD} are computed analogously. 
\item \textit{Assignment procedure:} For each non-elementary criterion $g_{\mathbf{r}}$,
and for each category $C_{h}$, a likeness threshold $\lambda_{\mathbf{r}}^{h}$
has to be defined. In this case, we are assuming that the likeness
thresholds are the same for each $g_{\mathbf{r}}$ and these values
are shown in Table \ref{MembershipDegree}.

\begin{table}[!htb]
\caption{Likeness threshold for the four categories}
\label{MembershipDegree} \centering \begin{small} \resizebox{0,38\textwidth}{!}{
\begin{tabular}{lllll}
\hline 
 & $h=1$  & $h=2$  & $h=3$  & $h=4$ \tabularnewline
\hline 
$\lambda_{\mathbf{r}}^{h}$  & 0.65  & 0.60  & 0.65  & 0.60 \tabularnewline
\hline 
\end{tabular}} \end{small} 
\end{table}

Comparing the partial likeness degree $\delta_{\mathbf{r}}(a_{3},\cdot)$
with the corresponding likeness threshold $\lambda_{\mathbf{r}}^{h}$
for each non-elementary criterion $g_{\mathbf{r}}$, soldier $a_{3}$
can be assigned to the categories shown in Table \ref{Assignments}.

\begin{table}[!htb]
\caption{Assignments of $a_{3}$ on each non-elementary criterion}
\label{Assignments}\centering\begin{small}\resizebox{0,9\textwidth}{!}{%
\begin{tabular}{c|lll|lll|lll|llc}
\hline 
 & $\delta_{\mathbf{1}}(a_{3},\cdot)$  & $\lambda_{\mathbf{1}}^{h}$  &  & $\delta_{\mathbf{2}}(a_{3},\cdot)$  & $\lambda_{\mathbf{2}}^{h}$  &  & $\delta_{\mathbf{3}}(a_{3},\cdot)$  & $\lambda_{\mathbf{3}}^{h}$  &  & $\delta_{\mathbf{0}}(a_{3},\cdot)$  & $\lambda_{\mathbf{0}}^{h}$  & \tabularnewline
\hline 
$b_{11}$  & 0.313  & 0.65  &  & 0.225  & 0.65  &  & 0  & 0.65  &  & 0  & 0.65  & \tabularnewline
$b_{21}$  & 0.856  & 0.60  & \checkmark  & 0.543  & 0.60  & \checkmark  & 0.668  & 0.60  & \checkmark  & 0.773  & 0.60  & \checkmark \tabularnewline
$b_{31}$  & 0  & 0.65  &  & 0  & 0.65  &  & 0  & 0.65  &  & 0  & 0.65  & \tabularnewline
$b_{41}$  & 1  & 0.60  & \checkmark  & 0.733  & 0.60  & \checkmark  & 0.926  & 0.60  & \checkmark  & 0.842  & 0.60  & \checkmark \tabularnewline
\hline 
\end{tabular}} \end{small} 
\end{table}
\end{enumerate}



\subsection{Application of the SMAA to the h\textsc{Cat-SD} method}

\noindent Considering the likeness thresholds for each category shown
in Table \ref{MembershipDegree}, and assuming that they are the same
for each non-elementary criterion $g_{\mathbf{r}}$, we applied the
h\textsc{Cat-SD} method for each sampled set of compatible parameters.
Therefore, we were able to compute the probability of assigning each
soldier to the considered categories reported in Table \ref{FrequenciesAssignments}.

\begin{table}[!h]
\centering{}\caption{Probability of assignments expressed in percentage}
\label{FrequenciesAssignments} \subtable[Comprehensive level\label{ComprehensiveFrequencies}]{\global\long\def\arraystretch{1}
 { \resizebox{0.5\textwidth}{!}{ %
\begin{tabular}{c|cccc|c|c}
\hline 
\textbf{Soldier}  & $C_{1}$  & $C_{2}$  & $C_{3}$  & $C_{4}$  & $\{C_{2},C_{4}\}$  & $C_{5}$\tabularnewline
\hline 
$a_{1}$  & 100  & 0  & 0  & 0  & 0  & 0 \tabularnewline
$a_{2}$  & 0  & 0  & 0  & 0  & 100  & 0 \tabularnewline
$a_{3}$  & 0  & 0  & 0  & 0  & 100  & 0 \tabularnewline
$a_{4}$  & 0  & 0  & 100  & 0  & 0  & 0 \tabularnewline
$a_{5}$  & 0  & 100  & 0  & 0  & 0  & 0 \tabularnewline
$a_{6}$  & 100  & 0  & 0  & 0  & 0  & 0 \tabularnewline
$a_{7}$  & 0  & 0  & 0  & 0  & 100  & 0 \tabularnewline
\hline 
\end{tabular}} } } \subtable[Mental Sharpness $(MS)$\label{MSFrequencies}]{\global\long\def\arraystretch{1}
 { \resizebox{0.5\textwidth}{!}{ %
\begin{tabular}{c|cccc|cc|c}
\hline 
\textbf{Soldier}  & $C_{1}$  & $C_{2}$  & $C_{3}$  & $C_{4}$  & $\{C_{1},C_{3}\}$  & $\{C_{2},C_{4}\}$  & $C_{5}$\tabularnewline
\hline 
$a_{1}$  & 0  & 0  & 0  & 0  & 100  & 0  & 0 \tabularnewline
$a_{2}$  & 0  & 100  & 0  & 0  & 0  & 0  & 0 \tabularnewline
$a_{3}$  & 0  & 0  & 0  & 0  & 0  & 100  & 0 \tabularnewline
$a_{4}$  & 0  & 0  & 0  & 0  & 0  & 0  & 100 \tabularnewline
$a_{5}$  & 0  & 0  & 0  & 0  & 0  & 100  & 0 \tabularnewline
$a_{6}$  & 100  & 0  & 0  & 0  & 0  & 0  & 0 \tabularnewline
$a_{7}$  & 0  & 0  & 0  & 0  & 0  & 100  & 0 \tabularnewline
\hline 
\end{tabular}} } } \subtable[Mental Resilience ($MR$)\label{MRFrequencies}]{\global\long\def\arraystretch{1}
 { \resizebox{0.5\textwidth}{!}{ %
\begin{tabular}{c|cccc|c|c|c}
\hline 
\textbf{Soldier}  & $C_{1}$  & $C_{2}$  & $C_{3}$  & $C_{4}$  & $\{C_{2},C_{4}\}$  & $\{C_{2},C_{3},C_{4}\}$  & $C_{5}$\tabularnewline
\hline 
$a_{1}$  & 100  & 0  & 0  & 0  & 0  & 0  & 0 \tabularnewline
$a_{2}$  & 0  & 0  & 0  & 0  & 100  & 0  & 0 \tabularnewline
$a_{3}$  & 0  & 0  & 0  & 0  & 100  & 0  & 0 \tabularnewline
$a_{4}$  & 0  & 0  & 0  & 0  & 0  & 100  & 0 \tabularnewline
$a_{5}$  & 0  & 0  & 0  & 0  & 100  & 0  & 0 \tabularnewline
$a_{6}$  & 100  & 0  & 0  & 0  & 0  & 0  & 0 \tabularnewline
$a_{7}$  & 0  & 0  & 0  & 0  & 100  & 0  & 0 \tabularnewline
\hline 
\end{tabular}} } } \subtable[Physical and other Features ($PoF$)\label{PoFFrequencies}]{\global\long\def\arraystretch{1}
 { \resizebox{0.5\textwidth}{!}{ %
\begin{tabular}{c|cccc|c|c}
\hline 
\textbf{Soldier}  & $C_{1}$  & $C_{2}$  & $C_{3}$  & $C_{4}$  & $\{C_{2},C_{4}\}$  & $C_{5}$\tabularnewline
\hline 
$a_{1}$  & 100  & 0  & 0  & 0  & 0  & 0 \tabularnewline
$a_{2}$  & 0  & 0  & 0  & 0  & 100  & 0 \tabularnewline
$a_{3}$  & 0  & 0  & 0  & 0  & 100  & 0 \tabularnewline
$a_{4}$  & 0  & 0  & 97.269  & 0  & 0  & 2.731 \tabularnewline
$a_{5}$  & 0  & 100  & 0  & 0  & 0  & 0 \tabularnewline
$a_{6}$  & 100  & 0  & 0  & 0  & 0  & 0 \tabularnewline
$a_{7}$  & 0  & 0  & 0  & 100  & 0  & 0 \tabularnewline
\hline 
\end{tabular}} } } 
\end{table}

Looking at Tables \ref{ComprehensiveFrequencies}-\ref{PoFFrequencies}
one can observe that the results are quite stable, that is, the frequency
of assignment is very close to 100\% in almost all cases. This is
due to the fact that the preference information provided by the DM
was quite precise and, consequently, the space of parameters compatible
with this information was quite narrow. However, one can observe the
following:
\begin{itemize}
\item[$-$] At comprehensive level (Table \ref{ComprehensiveFrequencies}), all
candidates are assigned to at least one category. In particular, $a_{1}$
and $a_{6}$ are surely suitable to be snipers ($C_{1}$), $a_{5}$
sure be assigned to the breachers ($C_{2}$), $a_{3}$ is surely suitable
to be a communication operator ($C_{3}$), while the other three candidates,
that is $a_{2}$, $a_{3}$ and $a_{7}$, can be indifferently included
among breachers or heavy weapons operators ($\{C_{2},C_{4}\}$);
\item[$-$] With respect to $MS$, only two candidates can be assigned with certainty
to a unique category. In particular, $a_{2}$ is always assigned to
breachers category ($C_{2}$) and $a_{6}$ is always assigned to snipers
$(C_{1})$; regarding the remaining candidates, $a_{1}$ can cover
indifferently both snipers and communications operators $(\{C_{1},C_{3}\})$,
$a_{3}$, $a_{5}$ and $a_{7}$ can be included in breachers and heavy
weapons operator categories simultaneously $(\{C_{2},C_{4}\})$; finally,
$a_{4}$ is not idoneous to any of the considered categories;
\item[$-$] On $MR$, all candidates are assigned with certainty to at least
one category. $a_{1}$ and $a_{6}$ are idoneous to be included in
the snipers category $(C_{1})$; $a_{4}$ has evaluations such that
he can be included indifferently in all categories apart from snipers
one $(\{C_{2},C_{3},C_{4}\})$; finally, all the other candidates
($a_{2}$, $a_{3}$, $a_{5}$ and $a_{7}$) can be breachers or heavy
weapons operators indifferently $(\{C_{2},C_{4}\})$;
\item[$-$] Considering $PoF$, there is a better distribution of the candidates
among the different categories: $a_{1}$ and $a_{6}$ are assigned
with certainty to the snipers $(C_{1})$; $a_{5}$ is surely assigned
to the breachers $(C_{2})$; $a_{4}$ is included among the communication
operators $(C_{3})$ with a frequency of the $97.269\%$, while he
is not assigned to any category in the remaining cases; $a_{7}$ is
certainly idoneous to be included in the heavy weapons operators category
$(C_{4})$. The remaining two candidates, that is $a_{2}$ and $a_{3}$
can be indifferently assigned to the breachers and heavy weapons operators
categories ($\{C_{2},C_{4}\}$).
\end{itemize}

\subsection{A deterministic nominal classification respecting some specified
requirements}


\noindent To conclude this section, we shall show how the classification
procedure described in Section \ref{sec:RobustAssignment} can be
applied to this problem to get a deterministic nominal classification
taking into account the results obtained by using the SMAA methodology
and the following additional requirements that are specified by the
DM for each non-elementary criterion $g_{\mathbf{r}}$: 
\begin{description}
\item [{R1)}] Each soldier should be assigned to a single category or to
the dummy one; 
\item [{R2)}] At least one soldier should be assigned to each $C_{h}$,
$h=1,\ldots,4$; 
\item [{R3)}] At most two soldiers should be assigned to each $C_{h}$,
$h=1,\ldots,4$; 
\item [{R4)}] At most two soldiers should not be assigned (at most two
soldiers should be assigned to the dummy category $C_{5}$). 
\end{description}
\noindent Taking into account the SMAA results given in tables \ref{ComprehensiveFrequencies}-\ref{PoFFrequencies},
one deterministic nominal classification can be obtained for each
non-elementary criterion. Anyway, in the following we shall explain
in detail how to get the deterministic nominal classification at comprehensive
level, that is considering $g_{\mathbf{0}}$.

Looking at Table \ref{ComprehensiveFrequencies} we observe that $a_{2}$,
$a_{3}$ and $a_{7}$ can be always simultaneously assigned to categories
$C_{2}$ and $C_{4}$. Therefore, since we would like to consider
a nominal classification assigning soldiers to one among $C_{1}-C_{4}$
or to the dummy category $C_{5}$ we rewrite the table \ref{ComprehensiveFrequencies}
as shown in Table \ref{NewFrequencies}.

\begin{table}[!htb]
\caption{Frequencies of assignments at comprehensive level}
\label{NewFrequencies}\centering\resizebox{0,4\textwidth}{!}{%
\begin{tabular}{c|cccc|c}
\hline 
\textbf{Soldier}  & $C_{1}$  & $C_{2}$  & $C_{3}$  & $C_{4}$  & $C_{5}$\tabularnewline
\hline 
$a_{1}$  & 100  & 0  & 0  & 0  & 0 \tabularnewline
$a_{2}$  & 0  & 100  & 0  & 100  & 0 \tabularnewline
$a_{3}$  & 0  & 100  & 0  & 100  & 0 \tabularnewline
$a_{4}$  & 0  & 0  & 100  & 0  & 0 \tabularnewline
$a_{5}$  & 0  & 100  & 0  & 0  & 0 \tabularnewline
$a_{6}$  & 100  & 0  & 0  & 0  & 0 \tabularnewline
$a_{7}$  & 0  & 100  & 0  & 100  & 0 \tabularnewline
\hline 
\end{tabular}} 
\end{table}

Considering that, in this case, $A=\{a_{1},\ldots,a_{7}\}$, a deterministic
nominal classification taking into account the probabilistic information
given by the SMAA methodology and the requirements provided by the
DM, one has to solve the following LP problem where all variables
are binary and constraints $[C1]-[C4]$ translate the requirements
provided by the DM:

\begin{equation}
\begin{array}{l}
\min L(\mathbf{y}_{\mathbf{0}})={\displaystyle \sum_{a\in A}\sum_{h=1}^{5}y_{a,\mathbf{0}}^{h}\sum_{k\neq h}b_{\mathbf{0}}^{k}(a),\;\;\mbox{subject to}}\\[2mm]
\left.\begin{array}{ll}
\mbox{for each \ensuremath{a\in A}}\;\;{\displaystyle \sum_{h=1}^{5}y_{a,\mathbf{0}}^{h}=1} & [C1]\\[2mm]
\mbox{for each \ensuremath{h=1,\ldots,4}}\;\;{\displaystyle \sum_{a\in A}y_{a,\mathbf{0}}^{h}\geqslant1} & [C2]\\[2mm]
\mbox{for each \ensuremath{h=1,\ldots,4}}\;\;{\displaystyle \sum_{a\in A}y_{a,\mathbf{0}}^{h}\leqslant2} & [C3]\\[2mm]
{\displaystyle \sum_{a\in A}y_{a,\mathbf{0}}^{5}\leqslant2} & [C4]\\[3mm]
y_{a,\mathbf{0}}^{h}\in\{0,1\},\;\forall a\in A,\;\forall h=1,\ldots,5.
\end{array}\right\} E^{LF}
\end{array}\label{LPLossFunction}
\end{equation}

Solving the LP (\ref{LPLossFunction}), we get $y_{1,\mathbf{0}}^{1,*}=y_{2,\mathbf{0}}^{4,*}=y_{3,\mathbf{0}}^{2,*}=y_{4,\mathbf{0}}^{3,*}=y_{5,\mathbf{0}}^{2,*}=y_{6,\mathbf{0}}^{1,*}=y_{7,\mathbf{0}}^{4,*}=1$,
while all the other binary variables are equal to zero. This means
that the deterministic nominal classification shown in the first column
of Table \ref{ClassificationsComprehensive} is therefore obtained.

\begin{table}[!htb]
\caption{Deterministic nominal classifications obtained at comprehensive level}
\label{ClassificationsComprehensive}\centering\resizebox{0,3\textwidth}{!}{%
\begin{tabular}{c|ccc}
\hline 
\textbf{Soldier}  & $1st$  & $2nd$  & $3rd$ \tabularnewline
\hline 
$a_{1}$  & $C_{1}$  & $C_{1}$  & $C_{1}$ \tabularnewline
$a_{2}$  & $C_{4}$  & $C_{2}$  & $C_{4}$ \tabularnewline
$a_{3}$  & $C_{2}$  & $C_{4}$  & $C_{4}$ \tabularnewline
$a_{4}$  & $C_{3}$  & $C_{1}$  & $C_{3}$ \tabularnewline
$a_{5}$  & $C_{2}$  & $C_{2}$  & $C_{2}$ \tabularnewline
$a_{6}$  & $C_{1}$  & $C_{1}$  & $C_{1}$ \tabularnewline
$a_{7}$  & $C_{4}$  & $C_{4}$  & $C_{2}$ \tabularnewline
\hline 
\end{tabular}} 
\end{table}

To check for the existence of another deterministic nominal classification,
considering that the optimal value of the loss function previously
found is $L_{\mathbf{y}_{\mathbf{0}}^{*}}=300$, one has to solve
the same LP problem (\ref{LPLossFunction}) with the addition of the
constraints

\[
\begin{array}{ll}
L(\mathbf{y}_{\mathbf{0}})=300 & [C1']\\[2mm]
y_{1,\mathbf{0}}^{1}+y_{2,\mathbf{0}}^{4}+y_{3,\mathbf{0}}^{2}+y_{4,\mathbf{0}}^{3}+y_{5,\mathbf{0}}^{2}+y_{6,\mathbf{0}}^{1}+y_{7,\mathbf{0}}^{4}\leqslant6 & [C2']\\[2mm]
\end{array}
\]

\noindent where $[C1']$ imposes that the optimal value of the loss
function should not be deteriorated, while $[C2']$ ensures that the
previous solutions of the problem is not found anymore. By proceeding
in this way, one gets $y_{1,\mathbf{0}}^{1,*}=y_{2,\mathbf{0}}^{2,*}=y_{3,\mathbf{0}}^{4,*}=y_{4,\mathbf{0}}^{3,*}=y_{5,\mathbf{0}}^{2,*}=y_{6,\mathbf{0}}^{1,*}=y_{7,\mathbf{0}}^{4,*}=1$
that provides the deterministic nominal classification shown in the
second column of Table \ref{ClassificationsComprehensive}. Proceeding
analogously, we find only another deterministic nominal classification
summarizing the results obtained by the application of the SMAA methodology
and compatible with the requirements provided by the DM that is shown
in the last column of Table \ref{ClassificationsComprehensive}.

A similar procedure can be used to obtain the deterministic nominal
classifications w.r.t. each of the three macro-criteria. We will not
give the detail of the computations in these cases but the obtained
classifications are shown in Tables \ref{MSClassifications}-\ref{PoFClassifications}.

\begin{table}[!h]
\centering{}\caption{Deterministic nominal classification at partial level}
\label{DeterministicClassifications} \subtable[Mental Sharpness $(MS)$\label{MSClassifications}]{\global\long\def\arraystretch{1}
 { \resizebox{0.3\textwidth}{!}{ %
\begin{tabular}{c|ccc}
\hline 
\textbf{Soldier}  & $1st$  & $2nd$  & $3rd$ \tabularnewline
\hline 
$a_{1}$  & $C_{3}$  & $C_{3}$  & $C_{3}$ \tabularnewline
$a_{2}$  & $C_{2}$  & $C_{2}$  & $C_{2}$ \tabularnewline
$a_{3}$  & $C_{2}$  & $C_{4}$  & $C_{4}$ \tabularnewline
$a_{4}$  & $C_{5}$  & $C_{5}$  & $C_{5}$ \tabularnewline
$a_{5}$  & $C_{4}$  & $C_{4}$  & $C_{2}$ \tabularnewline
$a_{6}$  & $C_{1}$  & $C_{1}$  & $C_{1}$ \tabularnewline
$a_{7}$  & $C_{4}$  & $C_{2}$  & $C_{4}$ \tabularnewline
\hline 
\end{tabular}} } } \subtable[Mental Resilience ($MR$)\label{MRClassifications}]{\global\long\def\arraystretch{1}
 { \resizebox{0.3\textwidth}{!}{ %
\begin{tabular}{c|cccccc}
\hline 
\textbf{Soldier}  & $1st$  & $2nd$  & $3rd$  & $4th$  & $5th$  & $6th$\tabularnewline
\hline 
$a_{1}$  & $C_{1}$  & $C_{1}$  & $C_{1}$  & $C_{1}$  & $C_{1}$  & $C_{1}$ \tabularnewline
$a_{2}$  & $C_{2}$  & $C_{4}$  & $C_{2}$  & $C_{4}$  & $C_{4}$  & $C_{2}$ \tabularnewline
$a_{3}$  & $C_{2}$  & $C_{2}$  & $C_{4}$  & $C_{4}$  & $C_{2}$  & $C_{4}$ \tabularnewline
$a_{4}$  & $C_{3}$  & $C_{3}$  & $C_{3}$  & $C_{3}$  & $C_{3}$  & $C_{3}$ \tabularnewline
$a_{5}$  & $C_{4}$  & $C_{4}$  & $C_{4}$  & $C_{2}$  & $C_{2}$  & $C_{2}$ \tabularnewline
$a_{6}$  & $C_{1}$  & $C_{1}$  & $C_{1}$  & $C_{1}$  & $C_{1}$  & $C_{1}$ \tabularnewline
$a_{7}$  & $C_{4}$  & $C_{2}$  & $C_{2}$  & $C_{2}$  & $C_{4}$  & $C_{4}$ \tabularnewline
\hline 
\end{tabular}} } } \subtable[Physical and other Features ($PoF$)\label{PoFClassifications}]{\global\long\def\arraystretch{1}
 { \resizebox{0.25\textwidth}{!}{ %
\begin{tabular}{c|cc}
\hline 
\textbf{Soldier}  & $1st$  & $2nd$ \tabularnewline
\hline 
$a_{1}$  & $C_{1}$  & $C_{1}$ \tabularnewline
$a_{2}$  & $C_{4}$  & $C_{2}$ \tabularnewline
$a_{3}$  & $C_{2}$  & $C_{4}$ \tabularnewline
$a_{4}$  & $C_{3}$  & $C_{3}$ \tabularnewline
$a_{5}$  & $C_{2}$  & $C_{2}$ \tabularnewline
$a_{6}$  & $C_{1}$  & $C_{1}$ \tabularnewline
$a_{7}$  & $C_{4}$  & $C_{4}$ \tabularnewline
\hline 
\end{tabular}} } } 
\end{table}


\section{Conclusions}


\noindent \label{Conclusions}In this paper, we proposed a comprehensive
method extending a recently proposed nominal classification, the \textsc{Cat-SD}
method. Firstly, we applied MCHP to the \textsc{Cat-SD} method. Thus,
we have introduced the hierarchical \textsc{Cat-SD}, h\textsc{Cat-SD}.
The hierarchical decomposition of a complex multiple criteria nominal
classification problem is then possible when applying \textsc{Cat-SD}.
Secondly, interactions and antagonistic effects between criteria structured
in a hierarchical way were handled in our method. Then, to elicit
the values of the criteria weights as well as the interactions and
antagonistic coefficients used in the h\textsc{Cat-SD} method, we
proposed a new development of the hierarchical and imprecise SRF method.
We applied SMAA to the h\textsc{Cat-SD} method with the aim of obtaining
the probablity with which an action is assigned to a category (or
categories) at a comprehensive level and at a macro-criterion level.
Finally, considering the concept of loss function, we proposed a procedure
that starting from the probabilistic assignments obtained by SMAA
provides a final classification that fulfills some requirements given
by the DM. Putting together all these aspects, we therefore built
the SMAA-h\textsc{Cat-SD} method. We presented a numerical example
to illustrate the application of SMAA-h\textsc{Cat-SD}.

The proposed method gives to the DM the possibility: 
\begin{itemize}
\item[$-$] To structure the set of criteria in a hierarchical way (logical subsets
of criteria can be created in the hierarchy); 
\item[$-$] To provide imprecise information for obtaining the criteria weights
as well as the interaction and antagonistic coefficients by using
the imprecise SRF method; 
\item[$-$] To analyze, for several sets of compatible parameters, the probability
of the assignment results provided by the \textsc{Cat-SD}, considering
all criteria or one macro-criterion only; 
\item[$-$] To obtain a final assignment that takes into account robustness concerns
as represented by the probabilistic classification provided by SMAA. 
\end{itemize}

Several advantages can be underlined with respect to the application
of the proposed method. The main features can be stated as follows: 
\begin{enumerate}
\item In situations in which the DM has to handle a great number of criteria
to assess actions, adopting h\textsc{Cat-SD} is a more adequate approach
than applying \textsc{Cat-SD} considering all criteria at the same
level; 
\item For the elicitation of the criteria weights and interaction and antagonistic
coefficients, it is easier for the DM thinking about a small number
of related criteria than a large number; 
\item Besides the possibility of eliciting criteria weights for subsets
of criteria, our method gives to the DM the possibility to provide
imprecise information during the process of determining them; 
\item Applying SMAA to the h\textsc{Cat-SD}, the DM can better understand
the decision problem at hand exploring the problem more in deep. 
\end{enumerate}

To sum up, in this work we have considered robustness concerns by
taking into account the set of all weights and interaction and antagonistic
coefficients compatible with preference information provided by the
DM, while taking advantage of the hierarchical structure of criteria.
Let us remark that: 
\begin{itemize}
\item[$-$] The extension of the SRF method to elicit weights of criteria as
well as interaction and antagonistic coefficients can be applied to
all \textsc{Electre} methods and, more in general, to all outranking
methods; 
\item[$-$] The procedure permitting to pass from the probabilistic classification
provided by SMAA to the final assignment can be applied to other probabilistic
versions of classification methods, also ordinal, such as \textsc{Electre
Tri} and its variants. 
\end{itemize}

Future research can rely on applying the SMAA-h\textsc{Cat-SD} method
to real-world nominal classification problems. Extending the method
to group decision making is also an interesting direction of research.
It could also be interesting to study procedures for aiding the elicitation
of preference information given by the DM to reduce the cognitive
effort required during the decision aiding process.

\section*{Acknowledgments}

\noindent 
This work was supported by national funds through Funda\c{c}{\~a}o para a Ci{\^e}ncia
 e a Tecnologia (FCT) with reference UID/CEC/50021/2019. Ana
Sara Costa acknowledges financial support from Universidade de Lisboa,
Instituto Superior T{'e}cnico, and CEG\nobreakdash-IST (PhD Scholarship).
Salvatore Corrente and Salvatore Greco wish to acknowledge the funding
by the FIR of the University of Catania ``BCAEA3, New developments
in Multiple Criteria Decision Aiding (MCDA) and their application
to territorial competitiveness\char`\"{} and by the research project
``Data analytics for entrepreneurial ecosystms, sustainable development
and wellbeing indices\char`\"{} of the Department of Economics and
Business of the University of Catania. Salvatore Greco has also benefited
of the fund ``Chance\char`\"{} of the University of Catania. Jos{'e}
Rui Figueira was partially supported under the Isambard Kingdom Brunel
Fellowship Scheme during a one-month stay (April-May 2018) at the
Portsmouth Business School (April-May 2018), U.K., and acknowledges
the support of the hSNS FCT Research Project (PTDC/EGE-OGE/30546/2017)
and the FCT grant SFRH/BSAB/139892/2018.

\noindent \newpage{}

\addcontentsline{toc}{section}{\numberline{}References} \bibliographystyle{model2-names}
\bibliography{Full_bibliography}

\end{document}